\pdfoutput=1

\documentclass[11pt]{article}

\usepackage[final]{acl}

\usepackage{times}
\usepackage{latexsym}
\usepackage{multirow}
\usepackage{adjustbox}
\usepackage[T1]{fontenc}

\usepackage[utf8]{inputenc}
\usepackage{amsmath} 

\usepackage{microtype}

\usepackage{inconsolata}

\usepackage{graphicx}
\graphicspath{ {./images/} }
\usepackage{subcaption}
\usepackage{xspace}
\usepackage{booktabs}
\usepackage{float}

\definecolor{lightred}{rgb}{1, 0.8, 0.8}  
\definecolor{lightgreen}{rgb}{0.77, 1, 0.77}  
\definecolor{sgq}{rgb}{0.0, 0.73, 1.0} 
\definecolor{query}{rgb}{0.2588, 0.7569, 0.5059} 
\definecolor{sg}{rgb}{0.7373, 0.5490, 0.1010} 
\definecolor{sa}{rgb}{0.7294, 0.3843, 0.2235} 

\definecolor{fa}{rgb}{0.7294, 0.3843, 0.2235} 

\newcommand{\figref}[1]{Figure~\ref{#1}\xspace}
\newcommand{\ourmethod}[0]{Step Guided Reasoning\xspace}

\newcommand{\our}[0]{SGR\xspace}
\newcommand{\sg}[0]{step guidance\xspace}
\newcommand{\sgq}[0]{step guidance question\xspace}
\newcommand{\sa}[0]{step answer\xspace}

\title{Step Guided Reasoning: Improving Mathematical Reasoning using Guidance Generation and Step Reasoning}

\author{
  Lang Cao$^\dagger$ \quad Yingtian Zou$^{\dagger2}$  \quad Chao Peng \quad Renhong Chen \quad Wu Ning \quad Yitong Li\thanks{~ corresponding author. \textbf{$^\dagger$} indicates equal contribution.} \\
  \small{\texttt{ \{caolang4,liyitong3\}@huawei.com}} \\
  Huawei Technologies Co., Ltd., China, $^2$Shanghai Jiaotong University\\
}

\begin{document}
\maketitle
\begin{abstract}
Mathematical reasoning has been challenging for large language models (LLMs), and the introduction of step-by-step Chain-of-Thought (CoT) inference has significantly advanced the mathematical capabilities of LLMs. However, current approaches either necessitate extensive inference datasets for training or depend on few-shot methods that frequently compromise computational accuracy. To address these fundamental limitations, we propose \ourmethod, a novel training-free adaptation framework that efficiently equips general-purpose pre-trained language models with enhanced mathematical reasoning capabilities. In this approach, LLMs reflect on small reasoning steps, similar to how humans deliberate and focus attention on what to do next. By incorporating this reflective process into the inference stage, LLMs can effectively guide their reasoning from one step to the next. Through extensive experiments, we demonstrate the significant effect of \ourmethod in enhancing mathematical performance in state-of-the-art language models -- Qwen2-72B-Instruct outperforms its math-specific counterpart, Qwen2.5-72B-Math-Instruct, on MMLU-STEM with a score of $90.9\%$, compared to $87.3\%$. The average scores of Qwen2-7B-Instruct and Qwen2-72B-Instruct increase from $27. 1\%$ to $36. 3\%$ and from $36. 5\%$ to $47. 4\%$ in the math domain, respectively.
\end{abstract}

\section{Introduction}
Since the introduction of Chain-of-Thought (CoT)~\cite{wei2022chain} reasoning on LLMs~\cite{naveed2024comprehensiveoverviewlargelanguage,zhao2023survey,vaswani2023attentionneed}, it has been demonstrated how reasoning abilities naturally emerge in sufficiently large language models through a simple technique called thought chaining prompts. This approach involves enriching the prompts~\cite{DBLP:journals/corr/abs-2402-07927} with thought chaining examples, which serve as demonstrations to guide the model's reasoning process. However, complex mathematical reasoning remains a significant challenge for LLMs~\cite{he2024olympiadbench}.
Even though the accuracy of LLMs in mathematical reasoning can be improved with the scaling of model parameters and that of the training data, the amount of high-quality CoT data~\cite{DBLP:conf/coling/ChengLZW24} becomes the bottleneck~\cite{hoffmann2022training}.

There are several approaches to tackle these challenges in the inference stage, and the methods discussed below significantly enhance the model's performance on both mathematical reasoning and MMLU-STEM benchmarks~\cite{hendrycks2021measuringmassivemultitasklanguage}.
Cumulative reasoning~\cite{zhang2023cumulative} has been proposed to make great improvements over MATH datasets~\cite{hendrycks2021measuring}. Cumulative reasoning significantly enhances problem-solving by decomposing the task into smaller, more manageable elements and builds upon prior propositions, improving the overall effectiveness of problem-solving.
Additionally, ~\citeauthor{zheng2023take} proposed a ``Take a Step Back'' prompt (SBP) method, which introduced overall concepts and principles to guide model reasoning using results from high-level descriptions of original questions.
Both of these schemes improve the accuracy of mathematical reasoning by generating intermediate but useful contexts, namely "scratchpad"~\cite{nye2021show}, during the inference phase.

\begin{figure*}[t!]
    \centering
    \includegraphics[width=0.75\textwidth]{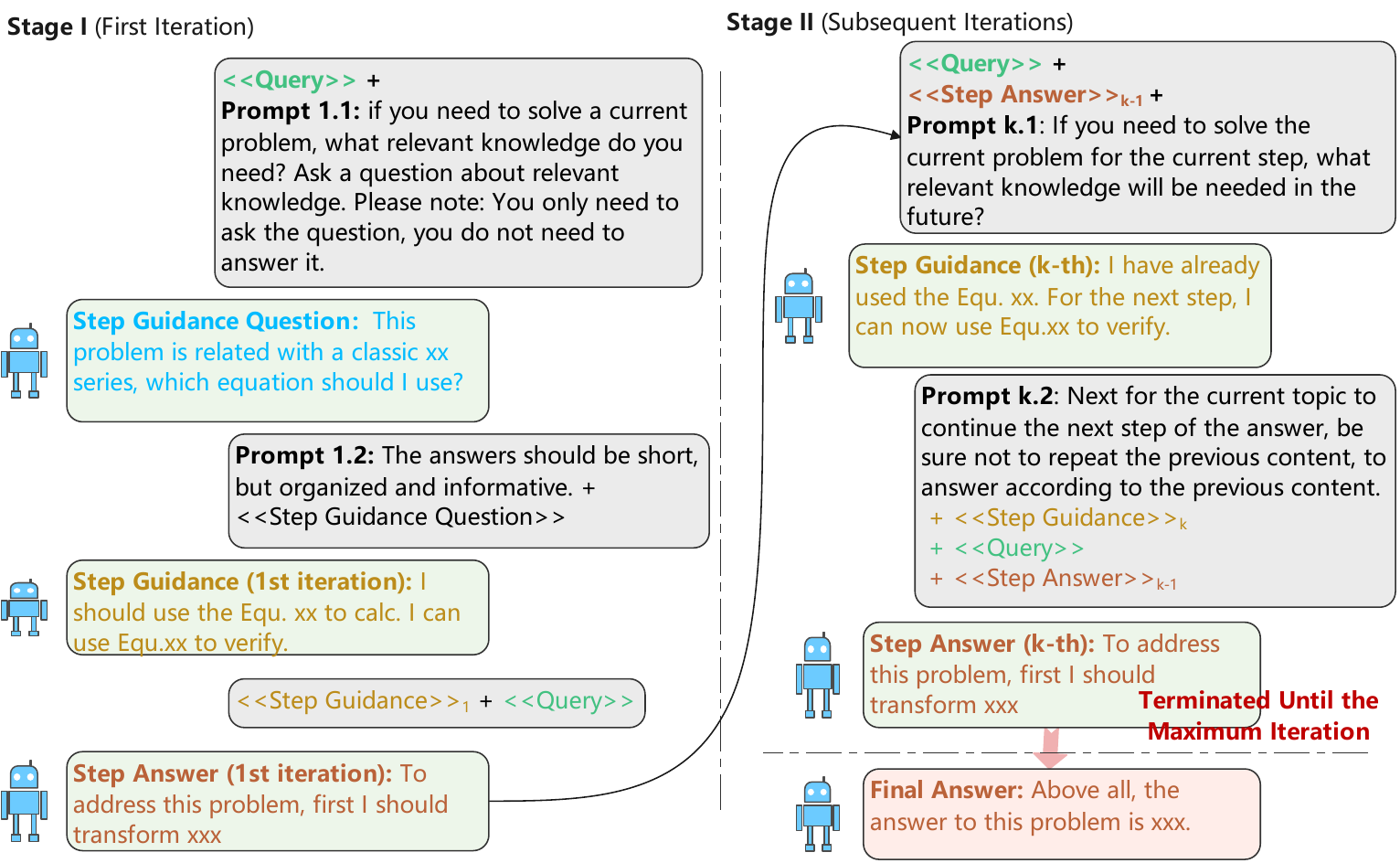}
    \caption{Illustration of how our proposed \our method generates \textcolor{sg}{step guidance} and \textcolor{sa}{step answer} for each iteration $k$. In stage I ($k=1$), \textbf{Prompt 1.1} questions the model to search for relevant knowledge. Subsequently, \textbf{Prompt 1.2} elicits a guidance from the model by getting it to answer the \textcolor{sgq}{\sgq}. Original \textcolor{query}{query} with such a \textcolor{sg}{step guidance} empowers the model to generate a more accurate and well-reasoned \textcolor{sa}{step answer}. In stage II ($1 < k\leq N$), the \textcolor{sa}{step answer} at step $k$ is refined by reiterating the process from \textcolor{sa}{step answer} $k-1$ with \textbf{Prompt k.1} and \textbf{k.2}. We iteratively enhance the \textcolor{sa}{step answer} until a satisfactory final answer is obtained. }

    \label{fig:overall_flow}
\end{figure*}

Another approach to enhancing mathematical reasoning ability involves methods that increase computation during the inference stage~\cite{zhang2024restmctsllmselftrainingprocess,DBLP:journals/corr/abs-2410-01707,yao2023treethoughtsdeliberateproblem,snell2024scalingllmtesttimecompute}. 
These approaches enable LLMs to explore multiple possible reasoning paths and select the most likely correct ones.
To be more specific, techniques such as Best-of-N (BoN)~\cite{sessa2024bondaligningllmsbestofn,DBLP:journals/tmlr/Dong0GZCPDZS023} and Tree-of-Thought (ToT)~\cite{yao2023treethoughtsdeliberateproblem} have also been explored. By scoring intermediate reasoning steps or evaluating the entire final result, the highest-scoring outcome by the reward model (RM)~\cite{DBLP:conf/nips/Ouyang0JAWMZASR22} is selected as the final answer. These strategies have been shown to effectively improve the model's mathematical reasoning ability, allowing it to tackle more complex problems with better accuracy and reliability.

Our observation in \figref{fig:step_MATH_amc} reveals that more challenging mathematical tasks often demand deeper and more deliberate multi-step reasoning. Motivated by this observation, we introduce \ourmethod (\our), a method that explicitly guides the model through step-by-step reasoning by encouraging more thoughtful intermediate steps. This is a simple yet highly flexible approach that dramatically enhances the reasoning abilities of general-purpose pretrained models without introducing any external knowledge. 
Unlike prior approaches that rely on an additional reward model, such as BoN, \textbf{\our can be seamlessly applied to any off-the-shelf pretrained model without fine-tuning, preserving its broad generalization abilities}. While its iterative design inevitably increases inference latency and token usage compared to single-pass prompting methods such as CoT or SBP, we show performance gain upon token costs in Figure \ref{fig:token}, which achieves higher accuracy but uses a smaller number of sampled tokens in total compared with resource-intensive strategies like Best-of-N sampling or R1-distilled models. Remarkably, when mathematical reasoning is required, 
\ourmethod can rapidly elevate a general LLM to expert-level performance, comparable to or even surpassing math-specific models or reasoning models. By applying our method, Qwen2-7B-Instruct improved the accuracy on the MATH dataset Level 5~\cite{hendrycks2021measuring}, the most difficult level, from $37.1\%$ to $58.6\%$, while Qwen2-Math-7B-Instruct achieved an accuracy of $52.0\%$. Similarly, Qwen2-72B-Instruct achieved an improvement from $35.8\%$ to $41.2\%$ on the OlympiaBench~\cite{he2024olympiadbench} open-ended, no-image English Math Competition test set, with Qwen2-Math-72B-Instruct achieving an accuracy of $42.5\%$. This characteristic makes \ourmethod an efficient and practical solution for deploying versatile LLMs as domain experts on demand, without sacrificing their general capabilities.

\section{Method}
\ourmethod (\our) method employs a series of reasoning steps during inference, each step consisting of generating two key components: a \textit{\sg} and a \textit{step answer}.\footnote{All used prompts are listed in Appendix~\ref{app:prompt}.}
The \textit{\sg} distills the most crucial logical elements and generates inferential cues. As a more sophisticated prompt signal, it fortifies every reasoning step. The \textit{step answer} then utilizes these cues comprehensively to produce more refined intermediate step responses. As a result, the overall reasoning becomes more efficient and impactful.

As illustrated in \figref{fig:overall_flow}, \our incorporates a multi-round iterative reasoning mechanism.
During the first iteration (Stage-I) of the reasoning, upon receiving a math query, we first direct the model to formulate a \textit{Step Guidance Question}. Subsequently, we prompt the model to engage in in-depth deliberation and response, thus eliciting a \textit{\sg}. 
This enables the model to generate a high-quality \textit{step answer} autonomously.
In the following iterative cycles (Stage II), we gradually leverage the step answer obtained from the preceding round to refine the step answer at the $k$-th step, until the model outputs a satisfactory result.

\our method provides a simple guidance mechanism that effectively promotes the model's thinking process for any potential auxiliary information, thereby fully exploiting the model’s inherent reasoning abilities. 
Inspired by CoT prompting, \our demonstrates that multi-step reasoning can be substantially enhanced through carefully designed self-guidance mechanisms alone. By iteratively decomposing complex mathematical problems into manageable sub-steps, our approach significantly improves both the accuracy and interpretability of the model’s reasoning process.

\subsection{Reasoning Step}
\our consists of multiple iterations as the \emph{reasoning steps} to instruct LLMs during inference. As shown in Figure \ref{fig:overall_flow}, the first step initiates a reasoning cycle (Stage I), and the subsequent steps (Stage II) iteratively refine the current step answer.
Each ``step'' can be defined at various granularities, including token-level~\cite{Zelikman2024QuietSTaRLM}, sentence-level~\cite{Jarrahi2023SLCNNsentencelevelconvolutionalneural}, paragraph-level~\cite{Chalkidis2021Paragraphlevelrationaleextractionregularization,DBLP:conf/asru/ZhangCLLW21}, or block-level, typically annotated by human experts~\cite{Lightman2023LetsVS}. In this paper, we opt to define a step as a paragraph level, since our approach focuses on challenging mathematical problems which generally require answers spanning thousands of tokens~\cite{Fu2023Complexitybasedpromptingmultistepreasoning}. Selecting appropriate granularity for the math domain ensures the effectiveness of instruction without losing coherence or logical flow while minimizing computational overhead.

In practice, we found delimiter ``\texttt{.\textbackslash n\textbackslash n}'' serves as an effective boundary for logical inference for most instruct models, such as GPT-4/GPT-4o, Qwen, and LLaMA. However, directly splitting reasoning at every occurrence of ``\texttt{.\textbackslash n\textbackslash n}'' can lead to repeated patterns in the model generation, causing the model to reanalyze the first step instead of progressing to the next. This issue arises because the model may interpret each split as a signal to reanalyze the problem, rather than advancing through the reasoning process.

To mitigate this problem, we introduce a step length constraint, where each step, delimited by ``\texttt{.\textbackslash n\textbackslash n}'', must contain a minimum number of characters. This helps ensure that each step contains sufficient information for meaningful reasoning and reduces the tendency for the model to repeat earlier analyses. 
Although this constraint addresses some of the repetition, LLMs could still exhibit long repetitive patterns in subsequent steps by chance, which would be fixed by fine-tuning to improve instruction following.

In theory, the step length required for different problems may vary, and even within a single problem, the length of steps may differ depending on the complexity of the reasoning required. Ideally, fine-tuning the language models over manually labelled data with a special step token could explicitly distinguish between steps, providing further clarity and precision in the reasoning process. However, this approach is not considered in the current paper, as our focus remains on leveraging an instruct-tuned model that requires no additional fine-tuning.

\subsubsection{Step Guidance}
For each iteration, the prompt guides the LLM to think about what relevant knowledge is needed next as \textit{\sg}, and the model is then asked to generate the corresponding reasoning as the \textit{step answer}.
The model does not revisit or retain \textit{previous step guidance}; instead, each generated \textit{\sg} is used exclusively for the current step, ensuring that each step is handled independently without carrying over unnecessary context.

For the first iteration, we adopt the SBP approach~\cite{zheng2023take} by using a question to obtain a more general \textit{\sg}. Specifically, in the first iteration, the model is prompted to independently generate a question related to the query as the \textit{\sgq}, and then the LLM answers this \textit{\sgq}, with the answer serving as the \textit{\sg}.

\subsubsection{Step Answer}
To generate the result of $k$-th reasoning step, both the generated \textit{\sg} at step $k$ and the previously accumulated $<<\text{step answer}>>_{k-1}$ are incorporated into the prompt to support continued reasoning. The generation process is halted once the model reaches the token ``\texttt{.\textbackslash n\textbackslash n}'' with a minimum length, which indicates the completion of the current step. This serves as a natural delimiter, ensuring that each step is sufficiently detailed and self-contained. To ensure the quality of generation of the $<<\text{step answer}>>_{k}$, we explicitly emphasized that "not to repeat the previous content" in the \textbf{Prompt k.2}. However, such repetitions still occurred. To address this, whenever a duplicate of the current step is detected, it is removed, and the model is prompted to resample and generate a new response. This trick ensures a streamlined reasoning process that eliminates unnecessary repetition, enabling the model to advance smoothly through each step without redundancy.

\subsection{Self-Reflection}
\our method mimics the human self-reflection process. It achieves this by iteratively refining the sub-steps required to reach a goal through multi-round internal self-questioning, which inherently constructs a chain of thought by itself. Compared to CoT, our iterative method does not merely break the reasoning process into multiple steps, but reflects any formulas and theorems that might be useful and evolve gradually. Therefore, \our encourages the model to reflect more from "its mind", thereby enhancing the quality of the rationale at each step within the chain of thought.

Distinct from Retrieval-Augmented Generation (RAG)~\cite{Gao2023Retrieval}, which leverages additional pre-existing or externally-retrieved context to enhance reasoning, our step answer mechanism hinges on step guidance where the additional context is excited by the model’s inherent reasoning abilities, rather than being sourced from external repositories. 
\ourmethod enables foundation models to attain competitive, contextually-aware multi-step reasoning performance on-the-fly without any need for task-specific fine-tuning or expensive test-time reflection process. It not only imparts greater flexibility and adaptability to the reasoning process, but also empowers foundation models to dynamically tailor their reasoning strategies, turns out to be a reasoning expert at hand.

\section{Experiments}

\begin{table*}[!t]
\centering
\resizebox{\linewidth}{!}{
\begin{tabular}{lcccccccccccc}
\toprule
\multicolumn{1}{c}{\multirow{2}{*}{}} &  \multicolumn{1}{c}{\multirow{2}{*}{Method}} & \multicolumn{6}{c}{\textbf{MATH}}                                 & \multicolumn{1}{c}{\multirow{2}{*}{\textbf{AMC23}}} & \multicolumn{1}{c}{\multirow{2}{*}{\textbf{AIME24}}}  & \multirow{2}{*}{\textbf{OLY}} & \multirow{2}{*}{\textbf{Average}} \\
\cmidrule{3-8}
 & & L1    & L2    & L3    & L4    & L5    & \multicolumn{1}{c}{Average} &  &  &  & \\
\midrule

\multicolumn{1}{l}{Qwen2-Math-72b-inst}      &  & \boxed{95.0} & \boxed{94.1} & \boxed{90.5} & \boxed{83.7} & \boxed{67.7} & \boxed{83.9}  & 60.0   & \boxed{20.0}  &  \boxed{42.5}   & \boxed{51.7}\\
\cmidrule{2-12}
\multirow{2}{*}{GPT-4o}   & CoT & \boxed{95.0} & 91.7 & 86.0 & 74.9 & 53.8 & 76.6 & 15.0 & 10.0 & 43.3 & 36.2 \\
 & SBP & 91.3 & 88.3 & 81.1 & 71.5 & 51.2 & 73.0 & 15.0 & 6.7 & 43.3 & 34.4\textcolor{gray}{(-1.8)} \\
\midrule
\multirow{3}{*}{Qwen2-72b-inst} 
&  CoT &  91.4 & 85.3 & 77.3 & 66.9 & 46.1 & 69.2 & 35.0 & 6.0 & 35.8 & 36.5                    \\
& SBP & 88.6 & 82.2 & 72.1 & 60.2 & 38.7 & 63.6 & 36.3 & 1.7 &32.7  & 33.6\textcolor{gray}{(-2.9)} \\
& L2M & 92.9 & 90.8 & 83.7 & 74.8 & 54.8 & 75.9 & 41.3
 & 6.7 & 44.0  & 42.0\textcolor{gray}{(+5.5)} \\
& \textbf{SGR}& \textbf{93.9} & \textbf{89.3} & \textbf{83.7} & \textbf{76.9} & \textbf{65.6} & \textbf{79.2}  & \boxed{\textbf{61.3}}    & \textbf{8.0} & \textbf{41.2} & \textbf{47.4}\textcolor{gray}{(+10.9)}                \\ \cmidrule{2-12}
\multirow{3}{*}{LLaMA3.1-8b-inst} 
&  CoT &  76.2 & 61.2 & 50.8 & 36.6 & 21.2 & 43.7                      & \textbf{20.0}    & \textbf{8.0}       & 14.4         & 21.5                \\
& SBP & 75.3 & 59.3 & 48.1 & 36.4 & 21.2 & 42.5 & 11.3 & 5.0 & 18.5 & 19.3\textcolor{gray}{(-2.2)} \\
&  L2M &  \textbf{85.2} & 72.4 & 62.4 & 48.7 & 31.6 & 54.7                      & 17.5    & 5.0       & \textbf{27.1}         &   26.1\textcolor{gray}{(+4.6)}             \\
& \textbf{SGR}&  81.7 & \textbf{76.8} & \textbf{71.5} & \textbf{66.8} & \textbf{61.2} & \textbf{69.5} & 18.8 & 6.0 & 22.7     & \textbf{29.2}\textcolor{gray}{(+7.7)}           
\\ 
\bottomrule
\end{tabular}
}
\end{table*}

\begin{table*}
\resizebox{\linewidth}{!}{

\begin{tabular}{lccccccccc}
\toprule
\multicolumn{1}{c}{\multirow{2}{*}{}} & \multicolumn{1}{c}{\multirow{2}{*}{Method}} & \multicolumn{7}{c}{\textbf{MMLU-STEM}}                                 \\
\cmidrule{3-9}
 &  & Physics    & Chemistry    & Biology    & Computer Science    & Math  & Engineer  & \multicolumn{1}{c}{Average}   \\
\midrule 
\multicolumn{1}{l}{Qwen2.5-Math-72b-inst}    &   & 88.2 & 78.7 & 86.9 & 83.9 & 92.6 & 81.2 & 87.3 
  \\
\cmidrule{2-9}
\multirow{2}{*}{GPT-4o}  & CoT & 90.0 & 64.8 & 94.7 & 85.3 & 87.8 & 83.3 & 86.1 \\
& SBP   & 89.6 & 82.1 & 95.1 & 87.0 & 87.9 & 77.8 & 87.8\textcolor{gray}{(+1.7)} \\
\midrule 
\multirow{3}{*}{Qwen2-72b-inst}  
& CoT & 86.3 & 74.9 & 93.8 & 81.8 & 86.5 & 75.3 & 85.3 \\
& SBP  & 81.8 & 70.6 & 91.4 & 80.3 & 82.7 & 71.9 & 81.5\textcolor{gray}{(-3.8)} \\
&L2M &80.8	& 71.9	& 89.7	& 82.8	& 86.5	& 76.8	& 83.0\textcolor{gray}{(-1.7)} \\
& \textbf{SGR}  & \boxed{\textbf{90.7}} & \boxed{\textbf{83.2}} & \boxed{\textbf{95.1}}& \boxed{\textbf{91.3}} & \boxed{\textbf{92.7}} & \textbf{78.8} & \boxed{\textbf{90.9}}\textcolor{gray}{(+5.6)}                  \\
\cmidrule{2-9}
\multirow{3}{*}{LLaMA3.1-8b-inst}  
& CoT & 59.4 & 62.4 & 56.1 & 78.4 & 61.2 & 64.9 &  69.2 \\
& SBP   & 62.7 &  57.7 & 77.6 &  60.2 & 65.4 & 65.7 &  64.9\textcolor{gray}{(-4.3)} \\
& L2M & 64.0 & 52.4 & 75.8 & 65.0 & 69.2 & 64.6 &  66.4\textcolor{gray}{(-2.8)} \\
& \textbf{SGR}  & \textbf{77.7} & \textbf{82.1} & \textbf{78.6} & \textbf{89.2} & \textbf{85.9} & \boxed{\textbf{81.1}} & \textbf{82.4}\textcolor{gray}{(+13.2)}     \\            
\bottomrule
\end{tabular}
}
\caption{Accuracy comparison (\%) of CoT, SBP(5-shot), Least-to-Most(L2M) and our \our methods with the SOTA over MATH (Level 1 to Level 5), AMC23, AIME24, MMLU-STEM and OLY datasets. We also compare the results of open-sourced SOTA math-specific models - the QwQ, Qwen-Math models and GPT-4o (full results refer to Table \ref{app:tab:comparison} in the Appendix). The best results of all are in \boxed{\text{Box}} and best results for each base are in \textbf{Bold}, and the grey numbers in the brackets indicate the improvements in terms of the models boosted by CoT.}
\label{tab:comparison}

\end{table*}

\subsection{Experimental Setups}
\paragraph{Datasets} 
For evaluation, we use four representative challenging math benchmarks, AMC23~\cite{aimovalidationamcdataset}, MATH~\cite{hendrycks2021measuring}, AIME24~\cite{aimovalidationdataset} and OlympiadBench (OLY)~\cite{he2024olympiadbenchchallengingbenchmarkpromoting}
with the open-ended, no-image English Math Competition (OE\_TO\_maths\_en\_COMP) tag. 
The selected mathematics test sets are all challenging and include competition-level questions (See Appendix~\ref{app:dataset}).

To assess the generalisability of our method, whether it is effective beyond mathematical reasoning domains, we selected MMLU~\cite{hendrycks2021measuringmassivemultitasklanguage} with STEM tags (MMLU-STEM) for evaluation. STEM, which encompasses the fields of Science, Technology, Engineering, and Mathematics, often requires specialized problem-solving skills. 
Each of the four datasets provides the problem as a query along with a reference answer, and we report the accuracy by comparing the final output of the LLM with the reference answer. 
Specifically, for the MMLU-STEM test dataset, a multiple-choice dataset, we determine the accuracy by comparing the final selected answer option with the reference answer. For the other test sets, we first accurately extract the final answer from the reference answer and then compare this extracted final answer with the answer generated by the model to ensure that the model output aligns with the intended task objectives. 
To ensure the reliability and consistency of our evaluation, we use GPT-4~\cite{openai2024gpt4technicalreport} as our validation tool, a model that has demonstrated near-human level evaluation capabilities~\cite{DBLP:conf/emnlp/SottanaLZY23}.

\paragraph{Models} 
Given that the \our method demands that LLMs display remarkably strong and comprehensive capabilities, we choose Qwen2-72B-Instruct, Qwen2-7B-Instruct~\cite{yang2024qwen2}, LLaMA3.1-8B-Instruct~\cite{grattafiori2024llama3herdmodels} and LLaMA2-70B-Instruct~\cite{touvron2023llama2openfoundation} as our experimental models.

We also use a distilled version of DeepSeek-R1 of Qwen-7b and LLaMA2-8b~\cite{deepseekai2025deepseekr1incentivizingreasoningcapability} to compare with Qwen-7b and LLaMA2-8B as the base instruct models promoted by our method. We also compared our method with the state-of-the-art expert models QwQ-32B-Preview~\cite{qwq-32b-preview}, Qwen2-Math-7B-Instruct, Qwen2-Math-72B-Instruct~\cite{yang2024qwen2}, Qwen2.5-Math-7B-Instruct, Qwen2.5-Math-72B-Instruct~\cite{yang2024qwen25mathtechnicalreportmathematical}, and GPT-4o~\cite{openai2023gpt4o}.

\paragraph{Comparisons}
Alongside the 0-shot CoT results for LLMs, we also compare with three representative methods: Best-of-N (BoN)~\cite{sessa2024bondaligningllmsbestofn}, ``Take a Step Back Prompt'' (SBP)~\cite{zheng2023take}, and Least-to-Most (L2M) prompting~\cite{zhou2023leasttomost}.
For BoN, we sample $16$ or $32$ responses for each problem using Qwen2-7B-Instruct and use Qwen2.5-Math-RM-72B~\cite{yang2024qwen25mathtechnicalreportmathematical} to score these responses, and select the highest score response as the final result.
For SBP, we adopt the original prompt template and examples from the SBP method, which is a 5-shot prompt to generate both the principal and the final answers.
For the L2M method, we follow the original implementation, prompting the model to decompose challenging problems into a sequence of simpler sub-questions and solve each sub-question.

\begin{table*}[!t]
\centering
\resizebox{0.74\linewidth}{!}{
\begin{adjustbox}{center}

\begin{tabular}{lccccccccc}
\toprule
\multicolumn{1}{c}{\multirow{2}{*}{}} & \multicolumn{1}{c}{\multirow{2}{*}{Method}} & \multicolumn{7}{c}{\textbf{MMLU-STEM}}                                 \\
\cmidrule{3-9}
 &  & Physics    & Chemistry    & Biology    & Computer Science    & Math  & Engineer  & \multicolumn{1}{c}{Average}   \\
  \midrule
  \midrule
\multirow{2}{*}{Qwen2-7b-inst} & CoT & 65.9 & 56.0 & 79.5 & 64.7 & 73.2 & 62.2 & 64.9\\
& \textbf{SGR}  & 79.2 & 72.3 &  \textbf{88.9} & \textbf{85.2} & 84.1 & \textbf{74.0} & \textbf{82.3}\textcolor{gray}{(+17.4)} \\
\midrule
DeepSeek-R1-Distill-Qwen-7b  & CoT & \textbf{81.0} & 75.1 & 71.7 & 72.4 & \textbf{90.8} & 72.2 & 80.6\textcolor{gray}{(+12.3)} \\
\midrule
\midrule
\multirow{2}{*}{LLaMA3.1-8b-inst}  & CoT & 59.4 & 62.4 & 56.1 & 78.4 & 61.2 & 64.9 &  69.2 \\
& \textbf{SGR}  & \textbf{77.7} & \textbf{82.1} & 78.6 & \textbf{89.2} & \textbf{85.9} & \textbf{81.1} & \textbf{82.4}\textcolor{gray}{(+13.2)}                 \\
\midrule
DeepSeek-R1-Distill-Llama-8b  & CoT & 74.9 & 75.1 & \textbf{81.2} & 70.7 & 82.5 & 65.3 & 77.2\textcolor{gray}{(+8.0)}        \\
\bottomrule
\end{tabular}

\end{adjustbox}
}
\caption{This figure compares the MMLU-STEM accuracy (\%) of LLaMA3.1-8B-series and Qwen2-7B-series under three conditions: (1) the Chain of Thought (CoT) results using the instruct model as baseline, (2) the results after applying the \our method through instruct models, and (3) the performance following distillation with DeepSeek-R1~\cite{deepseekai2025deepseekr1incentivizingreasoningcapability}. The best results of the same model are in \textbf{Bold}. }
\label{tab:r1}

\end{table*}

\paragraph{Hyperparamters} 
For the decoding strategy, we set temperature to $1.0$ and set top\_p to $1.0$ or $0.7$ for sampling.\footnote{We observed that top\_p decoding tends to mitigate repetition compared with greedy decoding.}
All experimental results are reported as the average accuracy scores under top\_p values of $0.7$ and $1.0$.
The step length constraint for MATH and MMLU-STEM was specified as $300$, while for the AIME24 dataset, it was set to $500$.
We use a maximum of 10 iterations for all test cases.
If there is a duplication between steps, it will delete and re-sample the solution in the current step.
We conducted the experiment using 8 V100 GPUs, with each problem in the test dataset generating an average output of 6,384 tokens from the MATH dataset by the Qwen2-7B-Instruct. We use float32 precision for the LLaMA3.1-8B-Instruct/Qwen2-7B-Instruct model, but float16 precision for the Qwen2-72B-Instruct model, leading to some degree of performance degradation. The native float16 precision is utilized for the LLaMA2-70B-Instruct model.

\subsection{Experimental Results}
The comparison results in Table \ref{tab:comparison} demonstrate the superior performance of our method (\our) across various datasets. On the MATH dataset, particularly with the Qwen2-72b-inst model, SGR achieves an average accuracy of $79.2\%$, marking a significant improvement over CoT's $69.2\%$.
This improvement is especially notable at the higher difficulty levels, where SGR demonstrates its robustness and effectiveness in handling complex problems. Similarly, with the LLaMA3.1-8b-inst model, SGR continues to outperform other methods across all levels, underscoring its adaptability and superior problem-solving capabilities. These results highlight the efficacy of SGR in improving model performance, making it a promising approach for complex computational tasks.

In Table \ref{tab:comparison}, SGR consistently demonstrates superior accuracy across six disciplines, extending beyond the math domain to showcase its effectiveness of general knowledge as well. Our method outperforms CoT with improvements of $5.6\%$ and $13.2\%$ on Qwen2-72b-inst and LLaMA3.1-8b-inst respectively. In contrast, SBP and L2M perform even lower than CoT, highlighting the substantial advantage SGR offers. These results underscore SGR's robust capability in enhancing model performance across diverse STEM disciplines, establishing it as a more effective approach compared to traditional methods. Full experimental results of other base models (QwQ-32b-Preview, Qwen2.5-Math-72b-inst, GPT4o, LLaMA2-70b-inst etc.) are shown in Appendix Table \ref{app:tab:comparison}.

\paragraph{Comparison to R1-Distilled Model} 
We also compare \our to reasoning-enhanced models which are distilled from DeepSeek-R1. In Table~\ref{tab:r1}, SGR significantly enhances the MMLU-STEM performance of base models, enabling them to surpass this distilled counterpart. Specifically, applying SGR to Qwen2-7b-inst boosted its average accuracy from $64.9\%$ (CoT) to $82.3\%$.
Similarly, the LLaMA3.1-8b-inst model, when augmented with SGR, outperforms DeepSeek-R1-Distill-Llama-8b and achieves high scores in Computer Science, Math, and Engineering. It is noteworthy that despite DeepSeek-R1 providing substantial reasoning knowledge to the base model, prompting it with traditional CoT still limits its reasoning ability when compared to the gains achieved by SGR. These results underscore SGR's substantial contribution to enhancing and unleashing the reasoning capabilities of instruct models on complex STEM tasks, positioning them favorably even against models specifically distilled for improved reasoning.

\subsection{Analysis}

\begin{figure}[t!]
    \centering
    \includegraphics[width=0.98\linewidth]{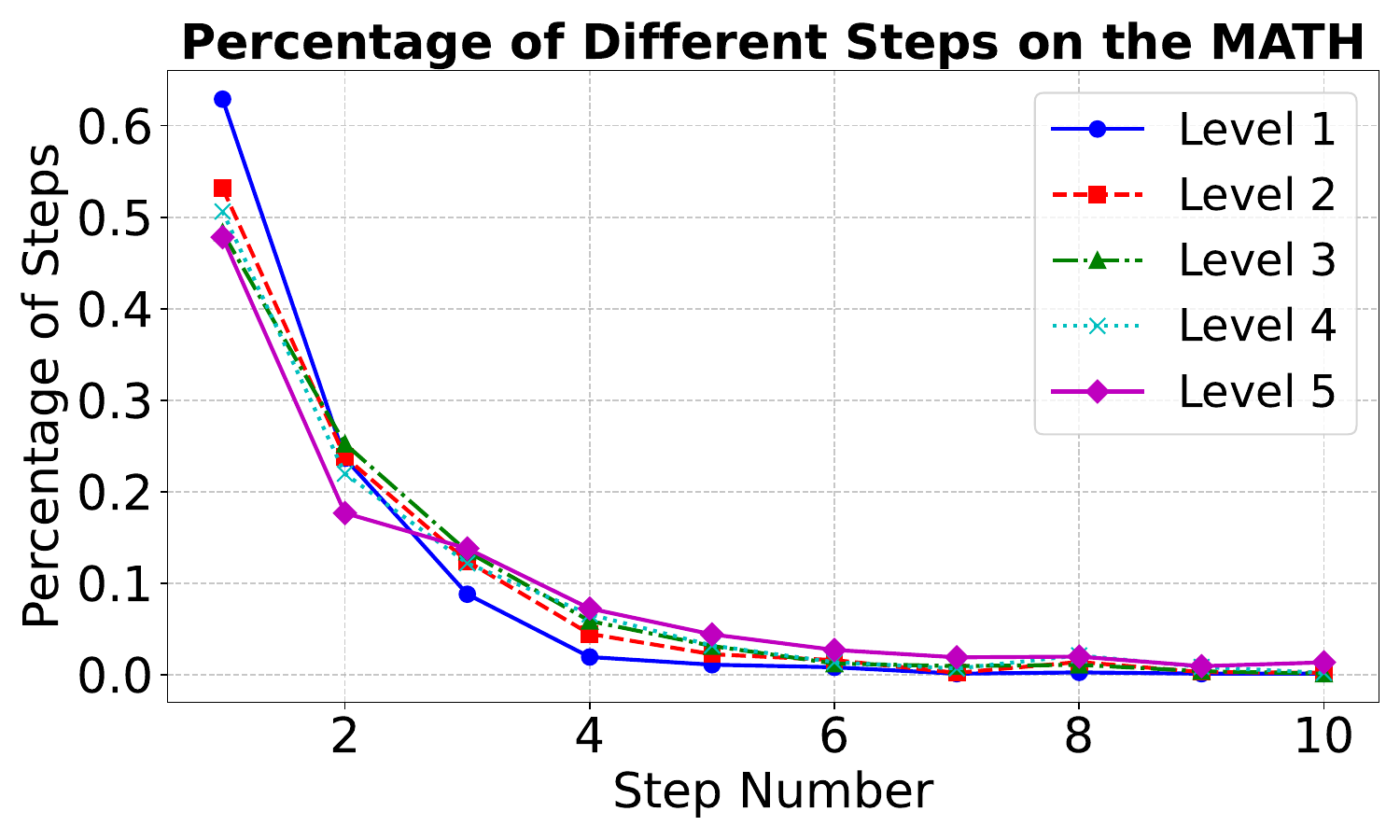}
    \caption{Illustrations of the proportion of different steps at where the correct answer first appears for problems across various difficulty levels over the MATH dataset. We report the average accuracies of the outputs from Qwen2-7b-Instruct with top\_p values of 0.7 and 1.0.}
    \label{fig:step_MATH_amc}
\end{figure}

\paragraph{Number of Reasoning Steps} We plot the number of steps required when the correct answer first appears in different levels on MATH. 
\figref{fig:step_MATH_amc} shows that the percentage of correct answers concentrates on the first four steps of reasoning (especially, $50\%$ of correct answers appear at the first step).
Intuitively, harder (higher-level) problems generally require more steps to reach a final solution compared to easier (lower-level) problems, which corroborates with the results.

\begin{figure}[!t]
    \centering
    \includegraphics[width=1.0\linewidth]{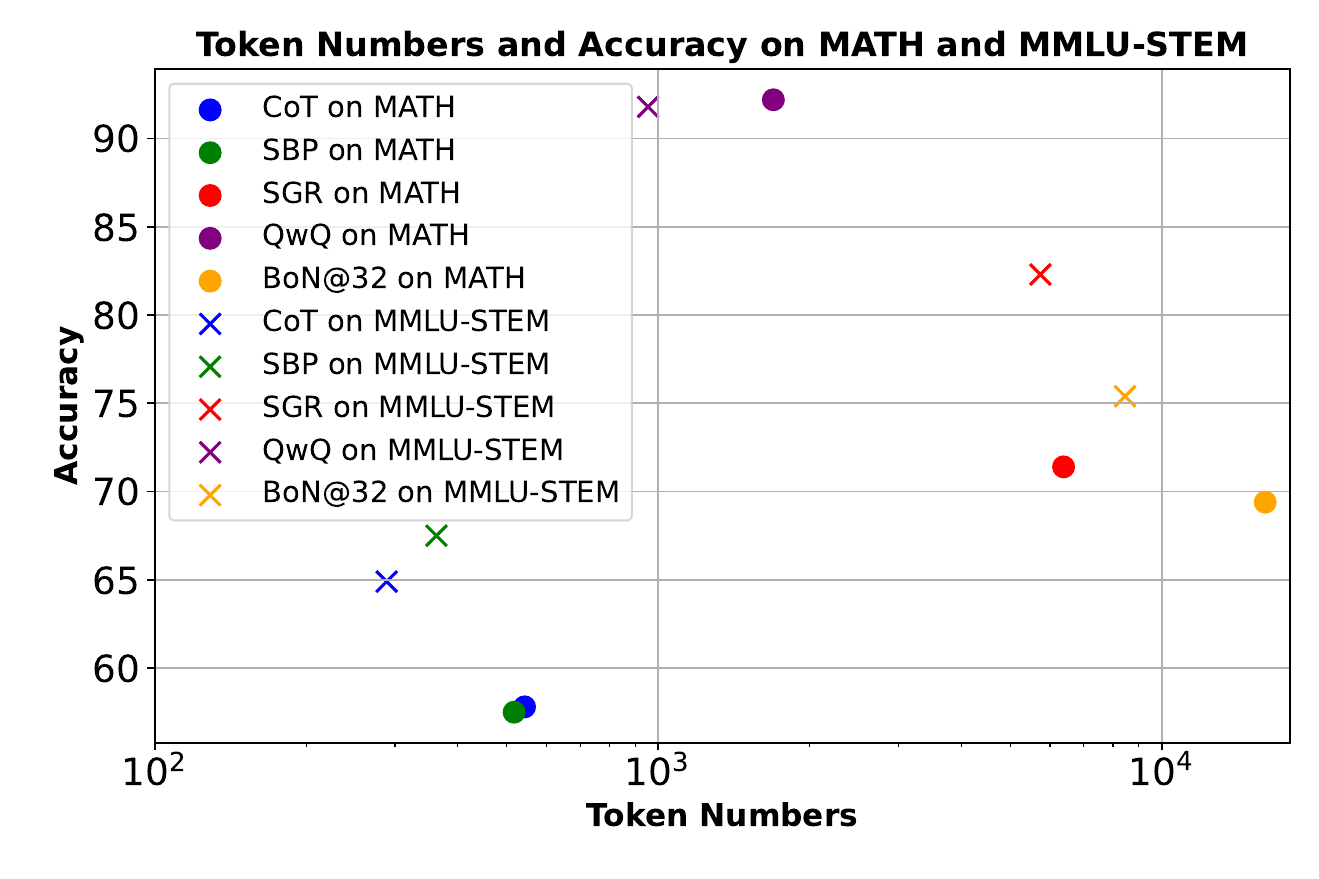}
    \caption{The scatter plot between the token numbers per query and accuracy for the MATH and MMLU-STEM datasets by Qwen2-7B-Instruct in different methods and QwQ-32B-Preview.}
    \label{fig:token}
\end{figure}

\paragraph{Token Numbers vs Accuracy}
\figref{fig:token} illustrates the relationship between the average number of tokens per query and the accuracy generated by the Qwen2-7B-Instruct model on the MATH and MMLU-STEM test sets using different methods. \our shows significantly higher results than CoT and SBP on both datasets. Notably, we achieve better results than BoN@32 while using less than half the number of tokens on MATH, which demonstrates the efficiency of our method.

\begin{figure}[t!]
    \centering
    \includegraphics[width=1.0\linewidth]{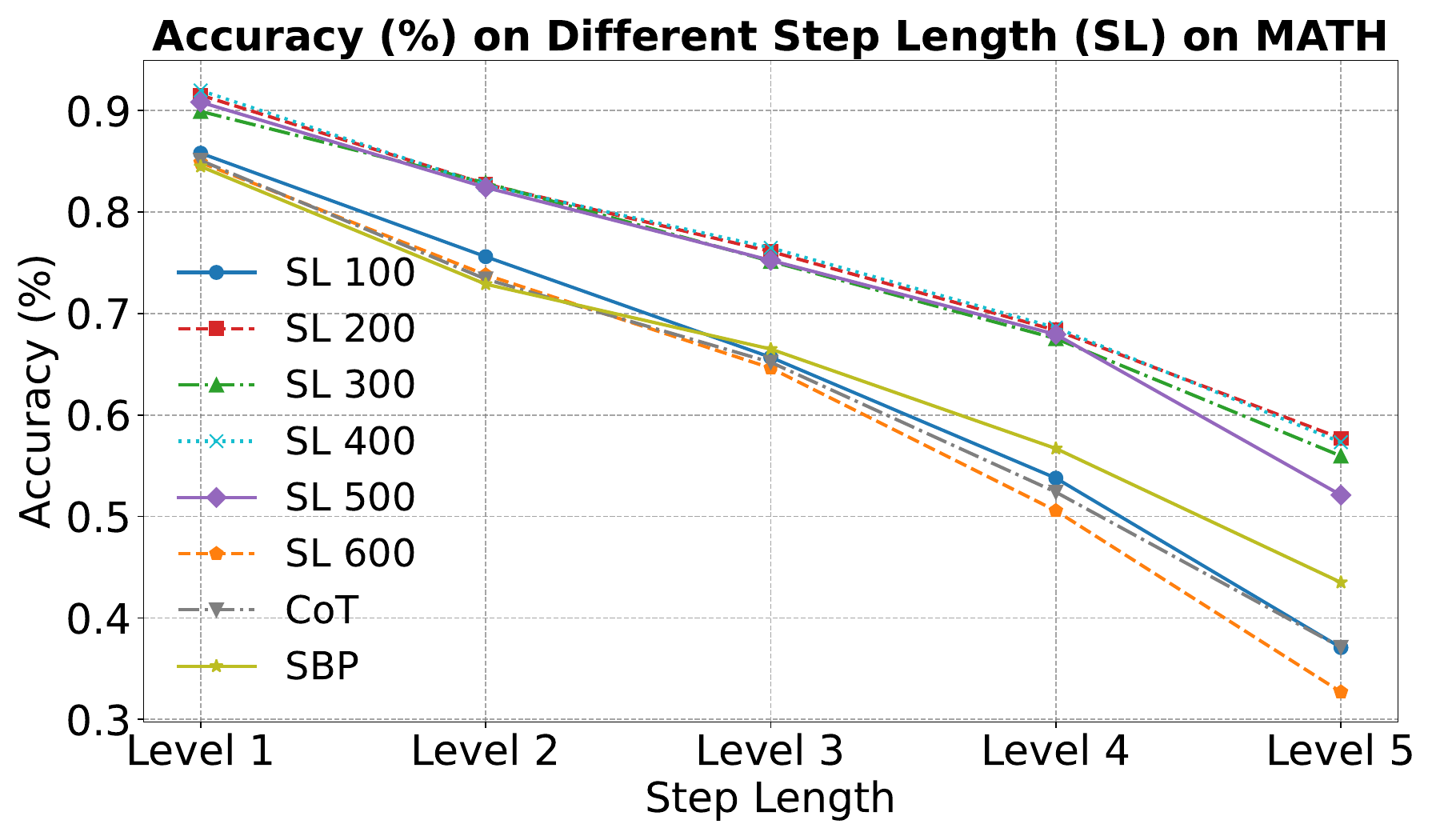}
    \caption{Accuracies vs. step length thresholds on the MATH dataset using \our over Qwen2-7B-Instruct. The 0-shot Chain of Thought (CoT) and Step-Back Prompt (SBP) generated by the same model are compared as the baseline.}
    \label{fig:step_number}
\end{figure}

\paragraph{Optimal Step Length}
We evaluate models' performance using different step lengths, ranging from $100$ to $600$, on the MATH dataset. The step length serves as a crucial hyperparameter, where the exact split point is dynamically determined by the first occurrence of the sequence ``\texttt{.\textbackslash n\textbackslash n}'' following the specified initial step length. 
As illustrated in \figref{fig:step_number}, we observe that when the step length ranges from $200$ to $500$, the accuracy is significantly higher compared to the baseline, with only minor variations in accuracy across this range. 

\paragraph{Case Study}
To understand how our method improves the reasoning procedure, we demonstrate an example in \figref{fig:sample}.
Compared to CoT at step 1, when calculating the ``second train'', the \sg generated by \our can help the model carry out the correct logical reasoning, while CoT reasoning makes an error.
Compared to CoT, our iterative method gives more guidance to the reasoning process.
The full contents of this example are included in the Appendix~\ref{app:sample}.

\begin{table*}[!t]
\centering
\resizebox{0.75\linewidth}{!}{
\begin{adjustbox}{center}

\begin{tabular}{lcccccccccccc}
\toprule
\multicolumn{1}{c}{\multirow{2}{*}{}} & \multicolumn{1}{c}{\multirow{2}{*}{Method}} & \multicolumn{6}{c}{\textbf{MATH}}                                 & \multicolumn{1}{c}{\multirow{2}{*}{\textbf{OLY}}} & \multirow{2}{*}{\textbf{AMC23}} & \multirow{2}{*}{\textbf{AIME24}} & \multirow{2}{*}{\textbf{Average}} \\
\cmidrule{3-8}
 &  & L1    & L2    & L3    & L4    & L5    & \multicolumn{1}{c}{Average} &  &  \\
\midrule
\multirow{4}{*}{Qwen2-7b-inst}  & CoT & 85.1 & 73.4 & 65.2 & 52.4 & 37.1 & 57.8      & 20.1  & 28.8 & 1.5  & 27.1               \\
& Stage I & \colorbox{lightred}{84.5} & \colorbox{lightred}{72.9} & \colorbox{lightgreen}{66.5} & \colorbox{lightgreen}{56.7} & \colorbox{lightgreen}{43.5} & \colorbox{lightgreen}{60.8} & \colorbox{lightgreen}{20.7}  & \colorbox{lightgreen}{32.5} & \colorbox{lightgreen}{\textbf{3.0}} & \colorbox{lightgreen}{29.3}\textcolor{gray}{(+2.2)} \\
& Stage II  & \colorbox{lightgreen}{88.6} & \colorbox{lightgreen}{77.7} & \colorbox{lightgreen}{68.8} & \colorbox{lightgreen}{58.4} & \colorbox{lightgreen}{40.1} & \colorbox{lightgreen}{62.3} & \colorbox{lightgreen}{\textbf{40.9}} & \colorbox{lightred}{27.5} & \colorbox{lightred}{0} & \colorbox{lightgreen}{32.7}\textcolor{gray}{(+5.6)} \\
& SGR & \colorbox{lightgreen}{90.2} & \colorbox{lightgreen}{81.3} & \colorbox{lightgreen}{74.6} & \colorbox{lightgreen}{68.3} & \colorbox{lightgreen}{\textbf{58.6}} & \colorbox{lightgreen}{\textbf{71.4}}  & \colorbox{lightgreen}{33.3} & \colorbox{lightgreen}{\textbf{38.8}} & \colorbox{lightred}{1.5} & \colorbox{lightgreen}{\textbf{36.3}}\textcolor{gray}{(+9.2)}              \\
 \midrule
 \multirow{4}{*}{Qwen2-72b-inst}  & CoT & 91.4 & 85.3 & 77.3 & 66.9 & 46.1 & 69.2 & 35.8   & 35.0 & 6.0 & 36.5 \\
& Stage I  & \colorbox{lightred}{88.1} & \colorbox{lightred}{80.4} & \colorbox{lightred}{74.1} & \colorbox{lightred}{62.7} & \colorbox{lightgreen}{46.8} & \colorbox{lightred}{66.5} & \colorbox{lightred}{31.6} &  \colorbox{lightgreen}{37.5} & \colorbox{lightred}{5.0} & \colorbox{lightred}{35.2}\textcolor{gray}{(-1.3)}                  \\
& Stage II  & \colorbox{lightgreen}{\textbf{93.9}} & \colorbox{lightgreen}{87.6} & \colorbox{lightgreen}{82.1} & \colorbox{lightgreen}{71.6} & \colorbox{lightgreen}{53.9} & \colorbox{lightgreen}{74.0}  & \colorbox{lightgreen}{\textbf{50.0}} & \colorbox{lightgreen}{45.0} & \colorbox{lightgreen}{6.7} & \colorbox{lightgreen}{43.9}\textcolor{gray}{(+7.4)}\\
 & SGR  & \colorbox{lightgreen}{\textbf{93.9}} & \colorbox{lightgreen}{\textbf{89.3}} & \colorbox{lightgreen}{\textbf{83.7}} & \colorbox{lightgreen}{\textbf{76.9}} & \colorbox{lightgreen}{\textbf{65.6}} & \colorbox{lightgreen}{\textbf{79.2}}   & \colorbox{lightgreen}{41.2}   & \colorbox{lightgreen}{\textbf{61.3}} & \colorbox{lightgreen}{\textbf{8.0}} & \colorbox{lightgreen}{\textbf{47.4}}\textcolor{gray}{(+10.9)}   \\
  \midrule
\multirow{4}{*}{LLaMA3.1-8b-inst}  & CoT & 76.2 & 61.2 & 50.8 & 36.6 & 21.2 & 43.7 & 14.4   & 20.0 & \textbf{8.0} & 21.5 \\
& Stage I  & \colorbox{lightred}{69.1} & \colorbox{lightred}{53.6} & \colorbox{lightred}{45.3} & \colorbox{lightred}{33.5} & \colorbox{lightgreen}{22.6} & \colorbox{lightred}{40.1}   & \colorbox{lightred}{12.6} & \colorbox{lightred}{18.8} & \colorbox{lightred}{\textbf{8.0}} & \colorbox{lightred}{19.9}\textcolor{gray}{(-1.6)}                  \\
& Stage II  & \colorbox{lightgreen}{77.6} & \colorbox{lightgreen}{66.0} & \colorbox{lightgreen}{55.6} & \colorbox{lightgreen}{43.5} & \colorbox{lightgreen}{27.6} & \colorbox{lightgreen}{49.1} & \colorbox{lightgreen}{\textbf{26.8}}  & \colorbox{lightgreen}{\textbf{23.8}} & \colorbox{lightred}{5.0} & \colorbox{lightgreen}{26.2}\textcolor{gray}{(+5.1)} \\
 & SGR  & \colorbox{lightgreen}{\textbf{81.7}} & \colorbox{lightgreen}{\textbf{76.8}} & \colorbox{lightgreen}{\textbf{71.5}} & \colorbox{lightgreen}{\textbf{66.8}} & \colorbox{lightgreen}{\textbf{61.2}} & \colorbox{lightgreen}{\textbf{69.5}} & \colorbox{lightgreen}{22.7} & \colorbox{lightred}{18.8} & \colorbox{lightred}{6.0} & \colorbox{lightgreen}{\textbf{29.3}}\textcolor{gray}{(+7.8)}   \\
  \midrule
\multirow{4}{*}{LLaMA2-70b-inst}  & CoT & \textbf{44.5} & 25.4 & 15.8 & 9.6 & 5.2 & 15.7 & 2.3 & 4.0 & 0.0 & 5.5 \\
& Stage I & \colorbox{lightred}{34.8} & \colorbox{lightred}{18.6} & \colorbox{lightred}{11.2} & \colorbox{lightred}{6.0} & \colorbox{lightred}{3.1} & \colorbox{lightred}{11.2} & \colorbox{lightgreen}{3.8} & \colorbox{lightred}{0.0} & \colorbox{lightgreen}{2.7} & \colorbox{lightred}{4.4}\textcolor{gray}{(-1.1)} \\
& Stage II & \colorbox{lightred}{43.6} & \colorbox{lightgreen}{\textbf{27.9}} & \colorbox{lightgreen}{\textbf{17.4}} & \colorbox{lightgreen}{\textbf{12.1}} & \colorbox{lightgreen}{\textbf{6.4}} & \colorbox{lightgreen}{\textbf{17.4}} & \colorbox{lightgreen}{\textbf{7.5}} & \colorbox{lightgreen}{\textbf{8.3}} & \colorbox{lightgreen}{\textbf{4.1}} & \colorbox{lightgreen}{\textbf{9.3}}\textcolor{gray}{(+3.8)} \\
 & SGR & \colorbox{lightred}{38.7} & \colorbox{lightred}{25.3} & \colorbox{lightgreen}{16.8} & \colorbox{lightgreen}{11.3} & \colorbox{lightgreen}{7.1} & \colorbox{lightgreen}{16.3} & \colorbox{lightgreen}{2.7} & \colorbox{lightgreen}{5.0} & \colorbox{lightgreen}{3.3} & \colorbox{lightgreen}{6.8}\textcolor{gray}{(+1.3)} \\
\bottomrule
\end{tabular}

\end{adjustbox}
}

\vspace{0.1cm}

\resizebox{0.75\linewidth}{!}{
\begin{adjustbox}{center}

\begin{tabular}{lccccccccc}
\toprule
\multicolumn{1}{c}{\multirow{2}{*}{}} & \multicolumn{1}{c}{\multirow{2}{*}{Method}} & \multicolumn{7}{c}{\textbf{MMLU-STEM}}                                 \\
\cmidrule{3-9}
 &  & Physics & Chemistry & Biology & Computer Science & Math  & Engineer & \multicolumn{1}{c}{Average}   \\
 \midrule
\multirow{4}{*}{Qwen2-7b-inst}  & CoT & 65.9 & 56.0 & 79.5 & 64.7 & 73.2 & 62.2
 & 64.9 \\
& Stage I  & \colorbox{lightred}{65.7} & \colorbox{lightred}{55.1} & \colorbox{lightred}{77.7} & \colorbox{lightgreen}{65.0} & \colorbox{lightred}{72.5} & \colorbox{lightred}{58.3} & \colorbox{lightred}{62.9}\textcolor{gray}{(-2.0)}\\
& Stage II  & \colorbox{lightgreen}{77.0} & \colorbox{lightgreen}{71.6} & \colorbox{lightgreen}{85.6} & \colorbox{lightgreen}{84.2} & \colorbox{lightgreen}{\textbf{84.9}} & \colorbox{lightgreen}{73.6} & \colorbox{lightgreen}{81.0}\textcolor{gray}{(16.1)}\\
& SGR  & \colorbox{lightgreen}{\textbf{79.2}} & \colorbox{lightgreen}{\textbf{72.3}} &  \colorbox{lightgreen}{\textbf{88.9}} & \colorbox{lightgreen}{\textbf{85.2}} & \colorbox{lightgreen}{84.1} & \colorbox{lightgreen}{\textbf{74.0}} & \colorbox{lightgreen}{\textbf{82.3}}\textcolor{gray}{(17.4)}              \\
\midrule
\multirow{4}{*}{Qwen2-72b-inst}  & CoT & 86.3 & 74.9 & 93.8 & 81.8 & 86.5 & 75.3 &  85.3 \\
& Stage I   & \colorbox{lightred}{84.8} & \colorbox{lightred}{71.1} & \colorbox{lightred}{90.7} & \colorbox{lightred}{79.8} & \colorbox{lightred}{83.6} & \colorbox{lightred}{70.5} & \colorbox{lightred}{82.9}\textcolor{gray}{(-2.4)} \\
& Stage II   & \colorbox{lightgreen}{\textbf{92.6}} & \colorbox{lightgreen}{\textbf{88.4}} & \colorbox{lightgreen}{\textbf{95.6}} & \colorbox{lightgreen}{\textbf{92.7}} & \colorbox{lightgreen}{92.0} & \colorbox{lightgreen}{\textbf{79.9}} & \colorbox{lightgreen}{\textbf{91.5}}\textcolor{gray}{(+6.2)} \\
& SGR  &  \colorbox{lightgreen}{90.7} & \colorbox{lightgreen}{83.2} & \colorbox{lightgreen}{95.1} & \colorbox{lightgreen}{91.3} & \colorbox{lightgreen}{\textbf{92.7}} & \colorbox{lightgreen}{78.8} & \colorbox{lightgreen}{90.9}\textcolor{gray}{(+5.6)}             \\
\midrule
\multirow{4}{*}{LLaMA3.1-8b-inst}  & CoT & 59.4 & 62.4 & 56.1 & 78.4 & 61.2 & 64.9 &  69.2 \\
& Stage I   & \colorbox{lightgreen}{59.7} & \colorbox{lightred}{61.4} & \colorbox{lightred}{54.0} & \colorbox{lightred}{77.0} & \colorbox{lightgreen}{62.0} & \colorbox{lightred}{60.9} & \colorbox{lightred}{67.9}\textcolor{gray}{(-1.3)} \\
& Stage II & \colorbox{lightgreen}{\textbf{82.8}} & \colorbox{lightgreen}{77.3} & \colorbox{lightgreen}{\textbf{91.7}} & \colorbox{lightgreen}{87.8} & \colorbox{lightgreen}{82.4} & \colorbox{lightgreen}{79.9} & \colorbox{lightgreen}{\textbf{83.7}}\textcolor{gray}{(+14.5)} \\
& SGR  & \colorbox{lightgreen}{77.7} & \colorbox{lightgreen}{\textbf{82.1}} & \colorbox{lightgreen}{78.6} & \colorbox{lightgreen}{\textbf{89.2}} & \colorbox{lightgreen}{\textbf{85.9}} & \colorbox{lightgreen}{\textbf{81.1}} & \colorbox{lightgreen}{82.4}\textcolor{gray}{(+13.2)}                 \\
\midrule
\multirow{4}{*}{LLaMA2-70b-inst}  & CoT  & 46.0 & 39.4 & 72.0 & 55.9 & 38.7 & 51.8 & 48.1 \\
& Stage I & \colorbox{lightgreen}{49.9} & \colorbox{lightgreen}{40.0} & \colorbox{lightgreen}{74.2} & \colorbox{lightred}{55.7} & \colorbox{lightred}{37.1} & \colorbox{lightgreen}{55.6} & \colorbox{lightgreen}{48.9}\textcolor{gray}{(+0.9)} \\
& Stage II & \colorbox{lightgreen}{\textbf{71.3}} & \colorbox{lightgreen}{\textbf{65.0}} & \colorbox{lightgreen}{\textbf{85.1}} & \colorbox{lightgreen}{\textbf{76.5}} & \colorbox{lightgreen}{\textbf{61.7}} & \colorbox{lightgreen}{\textbf{75.4}} & \colorbox{lightgreen}{\textbf{70.0}}\textcolor{gray}{(+21.9)} \\
& SGR  & \colorbox{lightgreen}{69.3} & \colorbox{lightgreen}{62.3} & \colorbox{lightgreen}{83.1} & \colorbox{lightgreen}{75.3} & \colorbox{lightgreen}{57.9} & \colorbox{lightgreen}{71.5} & \colorbox{lightgreen}{67.3}\textcolor{gray}{(+19.2)} \\
\bottomrule
\end{tabular}

\end{adjustbox}
}
\caption{Accuracy(\%) for Qwen2-7B-Instruct, Qwen2-72B-Instruct, LLaMA3.1-8B-Instruct and LLaMA2-70B-Instruct using different prompting methods on MATH, AMC23, AIME24, OLY and MMLU-STEM test datasets. The stage I refers to the initial iteration within \our framework (0-shot). The stage II is the second \our involves enhancing the first iteration by prompting the model from the outset to decide what action to take next. The best results are in \textbf{Bold} for each base. \colorbox{lightred}{Red} indicates lower results compared to the CoT baselines, while \colorbox{lightgreen}{Green} denotes higher results.}
\label{tab:ab}
\label{tab:ablation}
\end{table*}

\subsection{Ablation}

As shown in \figref{fig:overall_flow}, to explore the impact of each individual component, we extend ablation studies of the two stages of \our independently. In stage I, we prompt the LLMs to ask a \sgq without employing a few-shot template, allowing the model to answer the \sgq directly as the \sg. In stage II, we directly ask the model what knowledge it needs to use next and continue the process iteratively as \sg.

The experimental results are presented in Table~\ref{tab:ablation}.
When we check \sg and \sa, we find that for particularly challenging problems, e.g. OLY, the LLM struggles to generate the \sgq, often repeating the query. This severely undermines the effectiveness of \sg. However, when the LLM is allowed to directly use the prompt from Stage II to generate \sg, the quality of the \sg is significantly improved compared to Stage~I. 
As a result, \our achieves higher accuracy on OLY using only Stage II, outperforming the full \our.
However, we do not consider Stage I to be ineffective. This is because, compared with the complete SGR method, it can bring more significant improvements in the overall performance in MATH.

\begin{table}[t!]
\centering
\small
\begin{tabular}{@{}lcc@{}}
\toprule
\textbf{Models / Methods} & \textbf{SimpleQA} & \textbf{DROP} \\
\midrule
\textbf{Qwen3-8B-Instruct} & \multicolumn{1}{c}{--} & 79.5 \\
\textbf{QwQ-32B-Preview} & 2.0 & 74.1 \\
\textbf{Claude-3-Sonnet} & 5.1 & \multicolumn{1}{c}{--} \\
\textbf{GPT-3.5} & \multicolumn{1}{c}{--} & 64.1 \\
\midrule
\textbf{Qwen2-7B-instruct} & & \\
\addlinespace
\hspace{1em} \textit{0-shot CoT} & 2.4 & 59.9 \\
\hspace{1em} \textit{SBP} & 1.7 & 60.1 \\ 
\hspace{1em} \textit{L2M} & 1.6 & 62.8 \\
\hspace{1em} \textbf{\textit{SGR (Ours)}} & \textbf{2.6} & \textbf{70.7} \\
\midrule
\textbf{Qwen2-72B-instruct} & & \\
\addlinespace
\hspace{1em} \textit{0-shot CoT} & \textbf{6.9} & 68.6 \\
\hspace{1em} \textit{SBP} & 5.3 & 70.7 \\ 
\hspace{1em} \textit{L2M} & 4.7 & 59.9 \\
\hspace{1em} \textbf{\textit{SGR (Ours)}} & \textbf{6.9} & \textbf{75.6} \\
\bottomrule
\end{tabular}
\caption{Experiments on SimpleQA and DROP. All methods under each model section utilize the corresponding model, demonstrating SGR’s generalization capability across non-mathematical, factual and reasoning QA benchmarks.}
\label{tab:additional_experiments}
\end{table}

\section{Benchmarking on Diverse Non-STEM Tasks}

To assess the generalizability of our approach, we conducted extended experiments on two non-STEM datasets, SimpleQA~\cite{wei2024measuringshortformfactualitylarge} and DROP~\cite{dua2019dropreadingcomprehensionbenchmark}.
SimpleQA is a representative benchmark for factual QA and DROP is a complex reasoning QA task, respectively.
SimpleQA focuses on short, fact-based questions, while DROP requires multi-step reasoning and information synthesis over longer paragraphs.
Similarly, we compared our SGR method with several strong baselines, including SBP~\cite{zheng2023take}, L2M~\cite{zhou2023leasttomost}, and 0-shot CoT~\cite{wang2022self}.

Table~\ref{tab:additional_experiments} summarizes the results.
Notably, we observe that SGR consistently outperforms the baselines, especially on the DROP dataset, where DROP requires complex reasoning.
This highlights the effectiveness of \our in reasoning scenarios again and also demonstrates the generalization of our methods across non-mathematical reasoning QA benchmarks.

\section{Human Evaluation of SGR-Generated Answers on MATH}

To further verify the model's performance, we make human evaluations of two comparisons, SGR using Qwen2-72/7B-Instruct~\cite{yang2024qwen2} vs. a thinking model QwQ-32B-Preview~\cite{qwq-32b-preview}, and SGR vs. SBP.
We sampled $20$ questions from the MATH test set,  where we collected the corresponding answer pairs from both models for each sampled question.

For annotation, we invited $10$ mathematics annotators with extensive experience in mathematical data analysis to conduct a preference evaluation.
The evaluation was carried out double-blind that all annotators were unaware of which model produced each answer.
Each annotator independently assessed the answer pairs according to a standardized rubric, which included criteria such as correctness, clarity of reasoning, logical rigor, completeness of solution steps, and overall explanation quality.
Answers that demonstrated clear step-by-step reasoning, were well-organized, and provided sufficient explanation for each step were more likely to be favored.

\begin{figure}[!t]
    \centering
    \includegraphics[width=1.0\linewidth]{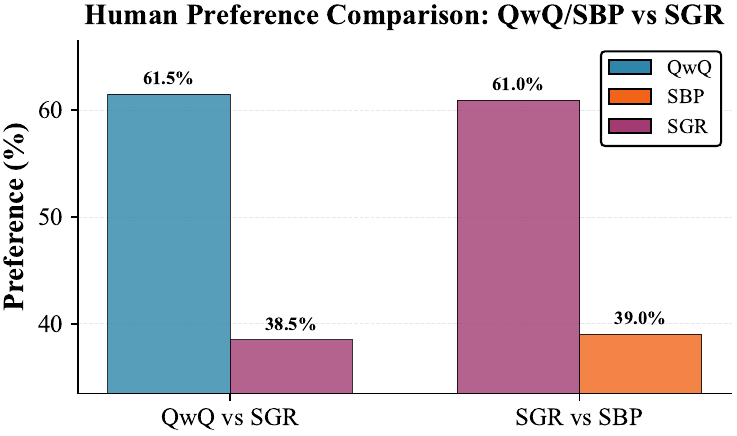}
    \caption{Human preference evaluation between QwQ-32B-Preview, Qwen2-7B/72B-Instruct models with SBP and Qwen2-7B/72B-Instruct models with SGR. Results show the percentage of cases where human annotators preferred each model's responses in pairwise comparisons across evaluation tasks. }
    \label{fig:survey}
\end{figure}

As shown in Figure~\ref{fig:survey}, QwQ-32B-Preview was favored in $61.5\%$ of cases, while the SGR method was preferred in $38.5\%$ of cases.
Although SGR did not outperform QwQ-32B-Preview, it still showed competitive performance, given that QwQ is specifically fine-tuned for thinking tasks while SGR is a tuning-free method.
When comparing with SBP, SGR method consistently outperformed SBP, and
Statistical analysis showed that the difference is significant.
These observations suggest that SGR can serve as a strong baseline or complementary approach, especially in scenarios where task-specific fine-tuning is not feasible.

\section{Conclusion}
We propose a step-by-step reasoning method that incorporates guidance generation within each step for multiple reasoning tasks.
Our method, applicable to general instruct LLMs without the need for further fine-tuning, employs self-questioning and self-answering at each reasoning step, where the model generates and answers to guide the step answer, enhancing the overall reasoning process.
When the model demonstrates a certain level of accuracy through CoT, it can significantly improve performance on challenging mathematical and logical reasoning problems across different-sized and series of models.
Compared with the SOTA methods, our approach can achieve stable improvements without the need for the Reward Model (RM), nor does it require fine-tuning.

\section*{Limitations}
Since extended reasoning sequences characteristic of our SGR approach demand considerable computational overhead, further optimization could unlock even greater performance gains from this large-scale model. Additionally, although we have verified that the SGR method leads to improvements across STEM domains, our evaluation has been primarily focused on mathematical reasoning tasks. The generalizability of our approach to more challenging AIGC tasks and broader domains remains an open question that warrants future investigation.

\bibliography{custom}

\begin{thebibliography}{43}
\providecommand{\natexlab}[1]{#1}

\bibitem[{AI-MO(2024{\natexlab{a}})}]{aimovalidationamcdataset}
AI-MO. 2024{\natexlab{a}}.
\newblock \href {https://huggingface.co/datasets/AI-MO/aimo-validation-amc} {Aimo validation amc dataset on hugging face}.

\bibitem[{AI-MO(2024{\natexlab{b}})}]{aimovalidationdataset}
AI-MO. 2024{\natexlab{b}}.
\newblock \href {https://huggingface.co/datasets/AI-MO/aimo-validation-aime} {Aimo validation dataset on hugging face}.

\bibitem[{Chalkidis et~al.(2021)Chalkidis, Fergadiotis, Tsarapatsanis, Aletras, Androutsopoulos, and Malakasiotis}]{Chalkidis2021Paragraphlevelrationaleextractionregularization}
Ilias Chalkidis, Manos Fergadiotis, Dimitrios Tsarapatsanis, Nikolaos Aletras, Ion Androutsopoulos, and Prodromos Malakasiotis. 2021.
\newblock \href {https://doi.org/10.18653/V1/2021.NAACL-MAIN.22} {Paragraph-level rationale extraction through regularization: {A} case study on european court of human rights cases}.
\newblock In \emph{Proceedings of the 2021 Conference of the North American Chapter of the Association for Computational Linguistics: Human Language Technologies, {NAACL-HLT} 2021, Online, June 6-11, 2021}, pages 226--241. Association for Computational Linguistics.

\bibitem[{Cheng et~al.(2024)Cheng, Li, Zhao, and Wen}]{DBLP:conf/coling/ChengLZW24}
Xiaoxue Cheng, Junyi Li, Wayne~Xin Zhao, and Ji{-}Rong Wen. 2024.
\newblock \href {https://aclanthology.org/2024.lrec-main.265} {Chainlm: Empowering large language models with improved chain-of-thought prompting}.
\newblock In \emph{Proceedings of the 2024 Joint International Conference on Computational Linguistics, Language Resources and Evaluation, {LREC/COLING} 2024, 20-25 May, 2024, Torino, Italy}, pages 2969--2983. {ELRA} and {ICCL}.

\bibitem[{Cobbe et~al.(2021)Cobbe, Kosaraju, Bavarian, Chen, Jun, Kaiser, Plappert, Tworek, Hilton, Nakano, Hesse, and Schulman}]{sessa2024bondaligningllmsbestofn}
Karl Cobbe, Vineet Kosaraju, Mohammad Bavarian, Mark Chen, Heewoo Jun, Lukasz Kaiser, Matthias Plappert, Jerry Tworek, Jacob Hilton, Reiichiro Nakano, Christopher Hesse, and John Schulman. 2021.
\newblock \href {https://arxiv.org/abs/2110.14168} {Training verifiers to solve math word problems}.
\newblock \emph{CoRR}, abs/2110.14168.

\bibitem[{DeepSeek-AI et~al.(2025)DeepSeek-AI, Guo, Yang, Zhang, Song, Zhang, Xu, Zhu, Ma, Wang, Bi, Zhang, Yu, Wu, Wu, and et~al.}]{deepseekai2025deepseekr1incentivizingreasoningcapability}
DeepSeek-AI, Daya Guo, Dejian Yang, Haowei Zhang, Junxiao Song, Ruoyu Zhang, Runxin Xu, Qihao Zhu, Shirong Ma, Peiyi Wang, Xiao Bi, Xiaokang Zhang, Xingkai Yu, Yu~Wu, Z.~F. Wu, and et~al. 2025.
\newblock \href {https://arxiv.org/abs/2501.12948} {Deepseek-r1: Incentivizing reasoning capability in llms via reinforcement learning}.
\newblock \emph{Preprint}, arXiv:2501.12948.

\bibitem[{Dong et~al.(2023)Dong, Xiong, Goyal, Zhang, Chow, Pan, Diao, Zhang, Shum, and Zhang}]{DBLP:journals/tmlr/Dong0GZCPDZS023}
Hanze Dong, Wei Xiong, Deepanshu Goyal, Yihan Zhang, Winnie Chow, Rui Pan, Shizhe Diao, Jipeng Zhang, Kashun Shum, and Tong Zhang. 2023.
\newblock \href {https://openreview.net/forum?id=m7p5O7zblY} {{RAFT:} reward ranked finetuning for generative foundation model alignment}.
\newblock \emph{Trans. Mach. Learn. Res.}, 2023.

\bibitem[{Dua et~al.(2019)Dua, Wang, Dasigi, Stanovsky, Singh, and Gardner}]{dua2019dropreadingcomprehensionbenchmark}
Dheeru Dua, Yizhong Wang, Pradeep Dasigi, Gabriel Stanovsky, Sameer Singh, and Matt Gardner. 2019.
\newblock \href {https://doi.org/10.18653/V1/N19-1246} {{DROP:} {A} reading comprehension benchmark requiring discrete reasoning over paragraphs}.
\newblock In \emph{Proceedings of the 2019 Conference of the North American Chapter of the Association for Computational Linguistics: Human Language Technologies, {NAACL-HLT} 2019, Minneapolis, MN, USA, June 2-7, 2019, Volume 1 (Long and Short Papers)}, pages 2368--2378. Association for Computational Linguistics.

\bibitem[{Dubey et~al.(2024)Dubey, Jauhri, Pandey, Kadian, Al{-}Dahle, Letman, Mathur, Schelten, Yang, Fan, Goyal, Hartshorn, Yang, Mitra, Sravankumar, Korenev, Hinsvark, Rao, Zhang, Rodriguez, Gregerson, Spataru, Rozi{\`{e}}re, Biron, Tang, and et~al.}]{grattafiori2024llama3herdmodels}
Abhimanyu Dubey, Abhinav Jauhri, Abhinav Pandey, Abhishek Kadian, Ahmad Al{-}Dahle, Aiesha Letman, Akhil Mathur, Alan Schelten, Amy Yang, Angela Fan, Anirudh Goyal, Anthony Hartshorn, Aobo Yang, Archi Mitra, Archie Sravankumar, Artem Korenev, Arthur Hinsvark, Arun Rao, Aston Zhang, Aur{\'{e}}lien Rodriguez, Austen Gregerson, Ava Spataru, Baptiste Rozi{\`{e}}re, Bethany Biron, Binh Tang, and et~al. 2024.
\newblock \href {https://doi.org/10.48550/ARXIV.2407.21783} {The llama 3 herd of models}.
\newblock \emph{CoRR}, abs/2407.21783.

\bibitem[{Fu et~al.(2023)Fu, Peng, Sabharwal, Clark, and Khot}]{Fu2023Complexitybasedpromptingmultistepreasoning}
Yao Fu, Hao Peng, Ashish Sabharwal, Peter Clark, and Tushar Khot. 2023.
\newblock \href {https://openreview.net/forum?id=yf1icZHC-l9} {Complexity-based prompting for multi-step reasoning}.
\newblock In \emph{The Eleventh International Conference on Learning Representations, {ICLR} 2023, Kigali, Rwanda, May 1-5, 2023}. OpenReview.net.

\bibitem[{Gao et~al.(2023)Gao, Xiong, Gao, Jia, Pan, Bi, Dai, Sun, Guo, Wang, and Wang}]{Gao2023Retrieval}
Yunfan Gao, Yun Xiong, Xinyu Gao, Kangxiang Jia, Jinliu Pan, Yuxi Bi, Yi~Dai, Jiawei Sun, Qianyu Guo, Meng Wang, and Haofen Wang. 2023.
\newblock \href {https://doi.org/10.48550/ARXIV.2312.10997} {Retrieval-augmented generation for large language models: {A} survey}.
\newblock \emph{CoRR}, abs/2312.10997.

\bibitem[{Gao et~al.(2024)Gao, Niu, He, Xu, Liu, Liu, Hu, and Wen}]{DBLP:journals/corr/abs-2410-01707}
Zitian Gao, Boye Niu, Xuzheng He, Haotian Xu, Hongzhang Liu, Aiwei Liu, Xuming Hu, and Lijie Wen. 2024.
\newblock \href {https://doi.org/10.48550/ARXIV.2410.01707} {Interpretable contrastive monte carlo tree search reasoning}.
\newblock \emph{CoRR}, abs/2410.01707.

\bibitem[{He et~al.(2024{\natexlab{a}})He, Luo, Bai, Hu, Thai, Shen, Hu, Han, Huang, Zhang, Liu, Qi, Liu, and Sun}]{he2024olympiadbench}
Chaoqun He, Renjie Luo, Yuzhuo Bai, Shengding Hu, Zhen~Leng Thai, Junhao Shen, Jinyi Hu, Xu~Han, Yujie Huang, Yuxiang Zhang, Jie Liu, Lei Qi, Zhiyuan Liu, and Maosong Sun. 2024{\natexlab{a}}.
\newblock \href {https://doi.org/10.18653/V1/2024.ACL-LONG.211} {Olympiadbench: {A} challenging benchmark for promoting {AGI} with olympiad-level bilingual multimodal scientific problems}.
\newblock In \emph{Proceedings of the 62nd Annual Meeting of the Association for Computational Linguistics (Volume 1: Long Papers), {ACL} 2024, Bangkok, Thailand, August 11-16, 2024}, pages 3828--3850. Association for Computational Linguistics.

\bibitem[{He et~al.(2024{\natexlab{b}})He, Luo, Bai, Hu, Thai, Shen, Hu, Han, Huang, Zhang, Liu, Qi, Liu, and Sun}]{he2024olympiadbenchchallengingbenchmarkpromoting}
Chaoqun He, Renjie Luo, Yuzhuo Bai, Shengding Hu, Zhen~Leng Thai, Junhao Shen, Jinyi Hu, Xu~Han, Yujie Huang, Yuxiang Zhang, Jie Liu, Lei Qi, Zhiyuan Liu, and Maosong Sun. 2024{\natexlab{b}}.
\newblock \href {https://doi.org/10.18653/V1/2024.ACL-LONG.211} {Olympiadbench: {A} challenging benchmark for promoting {AGI} with olympiad-level bilingual multimodal scientific problems}.
\newblock In \emph{Proceedings of the 62nd Annual Meeting of the Association for Computational Linguistics (Volume 1: Long Papers), {ACL} 2024, Bangkok, Thailand, August 11-16, 2024}, pages 3828--3850. Association for Computational Linguistics.

\bibitem[{Hendrycks et~al.(2021{\natexlab{a}})Hendrycks, Burns, Basart, Zou, Mazeika, Song, and Steinhardt}]{hendrycks2021measuringmassivemultitasklanguage}
Dan Hendrycks, Collin Burns, Steven Basart, Andy Zou, Mantas Mazeika, Dawn Song, and Jacob Steinhardt. 2021{\natexlab{a}}.
\newblock \href {https://openreview.net/forum?id=d7KBjmI3GmQ} {Measuring massive multitask language understanding}.
\newblock In \emph{9th International Conference on Learning Representations, {ICLR} 2021, Virtual Event, Austria, May 3-7, 2021}. OpenReview.net.

\bibitem[{Hendrycks et~al.(2021{\natexlab{b}})Hendrycks, Burns, Kadavath, Arora, Basart, Tang, Song, and Steinhardt}]{hendrycks2021measuring}
Dan Hendrycks, Collin Burns, Saurav Kadavath, Akul Arora, Steven Basart, Eric Tang, Dawn Song, and Jacob Steinhardt. 2021{\natexlab{b}}.
\newblock \href {https://datasets-benchmarks-proceedings.neurips.cc/paper/2021/hash/be83ab3ecd0db773eb2dc1b0a17836a1-Abstract-round2.html} {Measuring mathematical problem solving with the {MATH} dataset}.
\newblock In \emph{Proceedings of the Neural Information Processing Systems Track on Datasets and Benchmarks 1, NeurIPS Datasets and Benchmarks 2021, December 2021, virtual}.

\bibitem[{Hoffmann et~al.(2022)Hoffmann, Borgeaud, Mensch, Buchatskaya, Cai, Rutherford, de~Las~Casas, Hendricks, Welbl, Clark, Hennigan, Noland, Millican, van~den Driessche, Damoc, Guy, Osindero, Simonyan, Elsen, Rae, Vinyals, and Sifre}]{hoffmann2022training}
Jordan Hoffmann, Sebastian Borgeaud, Arthur Mensch, Elena Buchatskaya, Trevor Cai, Eliza Rutherford, Diego de~Las~Casas, Lisa~Anne Hendricks, Johannes Welbl, Aidan Clark, Tom Hennigan, Eric Noland, Katie Millican, George van~den Driessche, Bogdan Damoc, Aurelia Guy, Simon Osindero, Karen Simonyan, Erich Elsen, Jack~W. Rae, Oriol Vinyals, and Laurent Sifre. 2022.
\newblock \href {https://doi.org/10.48550/ARXIV.2203.15556} {Training compute-optimal large language models}.
\newblock \emph{CoRR}, abs/2203.15556.

\bibitem[{Jarrahi et~al.(2023)Jarrahi, Mousa, and Safari}]{Jarrahi2023SLCNNsentencelevelconvolutionalneural}
Ali Jarrahi, Ramin Mousa, and Leila Safari. 2023.
\newblock \href {https://doi.org/10.48550/ARXIV.2301.11696} {{SLCNN:} sentence-level convolutional neural network for text classification}.
\newblock \emph{CoRR}, abs/2301.11696.

\bibitem[{Lightman et~al.(2024)Lightman, Kosaraju, Burda, Edwards, Baker, Lee, Leike, Schulman, Sutskever, and Cobbe}]{Lightman2023LetsVS}
Hunter Lightman, Vineet Kosaraju, Yuri Burda, Harrison Edwards, Bowen Baker, Teddy Lee, Jan Leike, John Schulman, Ilya Sutskever, and Karl Cobbe. 2024.
\newblock \href {https://openreview.net/forum?id=v8L0pN6EOi} {Let's verify step by step}.
\newblock In \emph{The Twelfth International Conference on Learning Representations, {ICLR} 2024, Vienna, Austria, May 7-11, 2024}. OpenReview.net.

\bibitem[{Nye et~al.(2021)Nye, Andreassen, Gur{-}Ari, Michalewski, Austin, Bieber, Dohan, Lewkowycz, Bosma, Luan, Sutton, and Odena}]{nye2021show}
Maxwell~I. Nye, Anders~Johan Andreassen, Guy Gur{-}Ari, Henryk Michalewski, Jacob Austin, David Bieber, David Dohan, Aitor Lewkowycz, Maarten Bosma, David Luan, Charles Sutton, and Augustus Odena. 2021.
\newblock \href {https://arxiv.org/abs/2112.00114} {Show your work: Scratchpads for intermediate computation with language models}.
\newblock \emph{CoRR}, abs/2112.00114.

\bibitem[{OpenAI(2023)}]{openai2023gpt4o}
OpenAI. 2023.
\newblock Gpt-4o: Contributions.
\newblock \url{https://openai.com/gpt-4o-contributions/}.
\newblock Accessed: 2025-01-21.

\bibitem[{OpenAI et~al.(2024)OpenAI, Achiam, Adler, Agarwal, Ahmad, Akkaya, Aleman, Almeida, Altenschmidt, Altman, Anadkat, Avila, and et~al.}]{openai2024gpt4technicalreport}
OpenAI, Josh Achiam, Steven Adler, Sandhini Agarwal, Lama Ahmad, Ilge Akkaya, Florencia~Leoni Aleman, Diogo Almeida, Janko Altenschmidt, Sam Altman, Shyamal Anadkat, Red Avila, and et~al. 2024.
\newblock \href {https://arxiv.org/abs/2303.08774} {Gpt-4 technical report}.
\newblock \emph{Preprint}, arXiv:2303.08774.

\bibitem[{Ouyang et~al.(2022)Ouyang, Wu, Jiang, Almeida, Wainwright, Mishkin, Zhang, Agarwal, Slama, Ray, Schulman, Hilton, Kelton, Miller, Simens, Askell, Welinder, Christiano, Leike, and Lowe}]{DBLP:conf/nips/Ouyang0JAWMZASR22}
Long Ouyang, Jeffrey Wu, Xu~Jiang, Diogo Almeida, Carroll~L. Wainwright, Pamela Mishkin, Chong Zhang, Sandhini Agarwal, Katarina Slama, Alex Ray, John Schulman, Jacob Hilton, Fraser Kelton, Luke Miller, Maddie Simens, Amanda Askell, Peter Welinder, Paul~F. Christiano, Jan Leike, and Ryan Lowe. 2022.
\newblock \href {http://papers.nips.cc/paper\_files/paper/2022/hash/b1efde53be364a73914f58805a001731-Abstract-Conference.html} {Training language models to follow instructions with human feedback}.
\newblock In \emph{Advances in Neural Information Processing Systems 35: Annual Conference on Neural Information Processing Systems 2022, NeurIPS 2022, New Orleans, LA, USA, November 28 - December 9, 2022}.

\bibitem[{Sahoo et~al.(2024)Sahoo, Singh, Saha, Jain, Mondal, and Chadha}]{DBLP:journals/corr/abs-2402-07927}
Pranab Sahoo, Ayush~Kumar Singh, Sriparna Saha, Vinija Jain, Samrat Mondal, and Aman Chadha. 2024.
\newblock \href {https://doi.org/10.48550/ARXIV.2402.07927} {A systematic survey of prompt engineering in large language models: Techniques and applications}.
\newblock \emph{CoRR}, abs/2402.07927.

\bibitem[{Snell et~al.(2024)Snell, Lee, Xu, and Kumar}]{snell2024scalingllmtesttimecompute}
Charlie Snell, Jaehoon Lee, Kelvin Xu, and Aviral Kumar. 2024.
\newblock \href {https://doi.org/10.48550/ARXIV.2408.03314} {Scaling {LLM} test-time compute optimally can be more effective than scaling model parameters}.
\newblock \emph{CoRR}, abs/2408.03314.

\bibitem[{Sottana et~al.(2023)Sottana, Liang, Zou, and Yuan}]{DBLP:conf/emnlp/SottanaLZY23}
Andrea Sottana, Bin Liang, Kai Zou, and Zheng Yuan. 2023.
\newblock \href {https://doi.org/10.18653/V1/2023.EMNLP-MAIN.543} {Evaluation metrics in the era of {GPT-4:} reliably evaluating large language models on sequence to sequence tasks}.
\newblock In \emph{Proceedings of the 2023 Conference on Empirical Methods in Natural Language Processing, {EMNLP} 2023, Singapore, December 6-10, 2023}, pages 8776--8788. Association for Computational Linguistics.

\bibitem[{Team(2024)}]{qwq-32b-preview}
Qwen Team. 2024.
\newblock \href {https://qwenlm.github.io/blog/qwq-32b-preview/} {Qwq: Reflect deeply on the boundaries of the unknown}.

\bibitem[{Touvron et~al.(2023)Touvron, Martin, Stone, Albert, Almahairi, Babaei, Bashlykov, Batra, Bhargava, Bhosale, Bikel, Blecher, Canton{-}Ferrer, Chen, Cucurull, Esiobu, Fernandes, Fu, Fu, and et~al.}]{touvron2023llama2openfoundation}
Hugo Touvron, Louis Martin, Kevin Stone, Peter Albert, Amjad Almahairi, Yasmine Babaei, Nikolay Bashlykov, Soumya Batra, Prajjwal Bhargava, Shruti Bhosale, Dan Bikel, Lukas Blecher, Cristian Canton{-}Ferrer, Moya Chen, Guillem Cucurull, David Esiobu, Jude Fernandes, Jeremy Fu, Wenyin Fu, and et~al. 2023.
\newblock \href {https://doi.org/10.48550/ARXIV.2307.09288} {Llama 2: Open foundation and fine-tuned chat models}.
\newblock \emph{CoRR}, abs/2307.09288.

\bibitem[{Vaswani et~al.(2017)Vaswani, Shazeer, Parmar, Uszkoreit, Jones, Gomez, Kaiser, and Polosukhin}]{vaswani2023attentionneed}
Ashish Vaswani, Noam Shazeer, Niki Parmar, Jakob Uszkoreit, Llion Jones, Aidan~N. Gomez, Lukasz Kaiser, and Illia Polosukhin. 2017.
\newblock \href {https://proceedings.neurips.cc/paper/2017/hash/3f5ee243547dee91fbd053c1c4a845aa-Abstract.html} {Attention is all you need}.
\newblock In \emph{Advances in Neural Information Processing Systems 30: Annual Conference on Neural Information Processing Systems 2017, December 4-9, 2017, Long Beach, CA, {USA}}, pages 5998--6008.

\bibitem[{Wang et~al.(2023)Wang, Wei, Schuurmans, Le, Chi, Narang, Chowdhery, and Zhou}]{wang2022self}
Xuezhi Wang, Jason Wei, Dale Schuurmans, Quoc~V. Le, Ed~H. Chi, Sharan Narang, Aakanksha Chowdhery, and Denny Zhou. 2023.
\newblock \href {https://openreview.net/forum?id=1PL1NIMMrw} {Self-consistency improves chain of thought reasoning in language models}.
\newblock In \emph{The Eleventh International Conference on Learning Representations, {ICLR} 2023, Kigali, Rwanda, May 1-5, 2023}. OpenReview.net.

\bibitem[{Wei et~al.(2024)Wei, Karina, Chung, Jiao, Papay, Glaese, Schulman, and Fedus}]{wei2024measuringshortformfactualitylarge}
Jason Wei, Nguyen Karina, Hyung~Won Chung, Yunxin~Joy Jiao, Spencer Papay, Amelia Glaese, John Schulman, and William Fedus. 2024.
\newblock \href {https://doi.org/10.48550/ARXIV.2411.04368} {Measuring short-form factuality in large language models}.
\newblock \emph{CoRR}, abs/2411.04368.

\bibitem[{Wei et~al.(2022)Wei, Wang, Schuurmans, Bosma, Xia, Chi, Le, Zhou et~al.}]{wei2022chain}
Jason Wei, Xuezhi Wang, Dale Schuurmans, Maarten Bosma, Fei Xia, Ed~Chi, Quoc~V Le, Denny Zhou, et~al. 2022.
\newblock Chain-of-thought prompting elicits reasoning in large language models.
\newblock \emph{Advances in neural information processing systems}, 35:24824--24837.

\bibitem[{Yang et~al.(2024{\natexlab{a}})Yang, Yang, Hui, Zheng, Yu, Zhou, Li, Li, Liu, Huang, Dong, Wei, Lin, and et~al.}]{yang2024qwen2}
An~Yang, Baosong Yang, Binyuan Hui, Bo~Zheng, Bowen Yu, Chang Zhou, Chengpeng Li, Chengyuan Li, Dayiheng Liu, Fei Huang, Guanting Dong, Haoran Wei, Huan Lin, and et~al. 2024{\natexlab{a}}.
\newblock \href {https://doi.org/10.48550/ARXIV.2407.10671} {Qwen2 technical report}.
\newblock \emph{CoRR}, abs/2407.10671.

\bibitem[{Yang et~al.(2024{\natexlab{b}})Yang, Zhang, Hui, Gao, Yu, Li, Liu, Tu, Zhou, Lin, Lu, Xue, Lin, Liu, Ren, and Zhang}]{yang2024qwen25mathtechnicalreportmathematical}
An~Yang, Beichen Zhang, Binyuan Hui, Bofei Gao, Bowen Yu, Chengpeng Li, Dayiheng Liu, Jianhong Tu, Jingren Zhou, Junyang Lin, Keming Lu, Mingfeng Xue, Runji Lin, Tianyu Liu, Xingzhang Ren, and Zhenru Zhang. 2024{\natexlab{b}}.
\newblock \href {https://doi.org/10.48550/ARXIV.2409.12122} {Qwen2.5-math technical report: Toward mathematical expert model via self-improvement}.
\newblock \emph{CoRR}, abs/2409.12122.

\bibitem[{Yang et~al.(2024{\natexlab{c}})Yang, Xiang, Ge, Li, Lu, and Yu}]{naveed2024comprehensiveoverviewlargelanguage}
Haomiao Yang, Kunlan Xiang, Mengyu Ge, Hongwei Li, Rongxing Lu, and Shui Yu. 2024{\natexlab{c}}.
\newblock \href {https://doi.org/10.1109/MNET.2024.3367788} {A comprehensive overview of backdoor attacks in large language models within communication networks}.
\newblock \emph{{IEEE} Netw.}, 38(6):211--218.

\bibitem[{Yao et~al.(2024)Yao, Yu, Zhao, Shafran, Griffiths, Cao, and Narasimhan}]{yao2023treethoughtsdeliberateproblem}
Shunyu Yao, Dian Yu, Jeffrey Zhao, Izhak Shafran, Tom Griffiths, Yuan Cao, and Karthik Narasimhan. 2024.
\newblock Tree of thoughts: Deliberate problem solving with large language models.
\newblock \emph{Advances in Neural Information Processing Systems}, 36.

\bibitem[{Zelikman et~al.(2024)Zelikman, Harik, Shao, Jayasiri, Haber, and Goodman}]{Zelikman2024QuietSTaRLM}
Eric Zelikman, Georges Harik, Yijia Shao, Varuna Jayasiri, Nick Haber, and Noah~D. Goodman. 2024.
\newblock \href {https://doi.org/10.48550/ARXIV.2403.09629} {Quiet-star: Language models can teach themselves to think before speaking}.
\newblock \emph{CoRR}, abs/2403.09629.

\bibitem[{Zhang et~al.(2024)Zhang, Zhoubian, Yue, Dong, and Tang}]{zhang2024restmctsllmselftrainingprocess}
Dan Zhang, Sining Zhoubian, Yisong Yue, Yuxiao Dong, and Jie Tang. 2024.
\newblock \href {https://doi.org/10.48550/ARXIV.2406.03816} {Rest-mcts*: {LLM} self-training via process reward guided tree search}.
\newblock \emph{CoRR}, abs/2406.03816.

\bibitem[{Zhang et~al.(2021)Zhang, Chen, Li, Liu, and Wang}]{DBLP:conf/asru/ZhangCLLW21}
Qinglin Zhang, Qian Chen, Yali Li, Jiaqing Liu, and Wen Wang. 2021.
\newblock \href {https://doi.org/10.1109/ASRU51503.2021.9688078} {Sequence model with self-adaptive sliding window for efficient spoken document segmentation}.
\newblock In \emph{{IEEE} Automatic Speech Recognition and Understanding Workshop, {ASRU} 2021, Cartagena, Colombia, December 13-17, 2021}, pages 411--418. {IEEE}.

\bibitem[{Zhang et~al.(2023)Zhang, Yang, Yuan, and Yao}]{zhang2023cumulative}
Yifan Zhang, Jingqin Yang, Yang Yuan, and Andrew~Chi{-}Chih Yao. 2023.
\newblock \href {https://doi.org/10.48550/ARXIV.2308.04371} {Cumulative reasoning with large language models}.
\newblock \emph{CoRR}, abs/2308.04371.

\bibitem[{Zhao et~al.(2023)Zhao, Zhou, Li, Tang, Wang, Hou, Min, Zhang, Zhang, Dong, Du, Yang, Chen, Chen, Jiang, Ren, Li, Tang, Liu, Liu, Nie, and Wen}]{zhao2023survey}
Wayne~Xin Zhao, Kun Zhou, Junyi Li, Tianyi Tang, Xiaolei Wang, Yupeng Hou, Yingqian Min, Beichen Zhang, Junjie Zhang, Zican Dong, Yifan Du, Chen Yang, Yushuo Chen, Zhipeng Chen, Jinhao Jiang, Ruiyang Ren, Yifan Li, Xinyu Tang, Zikang Liu, Peiyu Liu, Jian{-}Yun Nie, and Ji{-}Rong Wen. 2023.
\newblock \href {https://doi.org/10.48550/ARXIV.2303.18223} {A survey of large language models}.
\newblock \emph{CoRR}, abs/2303.18223.

\bibitem[{Zheng et~al.(2024)Zheng, Mishra, Chen, Cheng, Chi, Le, and Zhou}]{zheng2023take}
Huaixiu~Steven Zheng, Swaroop Mishra, Xinyun Chen, Heng-Tze Cheng, Ed~H. Chi, Quoc~V Le, and Denny Zhou. 2024.
\newblock \href {https://openreview.net/forum?id=3bq3jsvcQ1} {Take a step back: Evoking reasoning via abstraction in large language models}.
\newblock In \emph{The Twelfth International Conference on Learning Representations}.

\bibitem[{Zhou et~al.(2023)Zhou, Sch{\"a}rli, Hou, Wei, Scales, Wang, Schuurmans, Cui, Bousquet, Le, and Chi}]{zhou2023leasttomost}
Denny Zhou, Nathanael Sch{\"a}rli, Le~Hou, Jason Wei, Nathan Scales, Xuezhi Wang, Dale Schuurmans, Claire Cui, Olivier Bousquet, Quoc~V Le, and Ed~H. Chi. 2023.
\newblock \href {https://openreview.net/forum?id=WZH7099tgfM} {Least-to-most prompting enables complex reasoning in large language models}.
\newblock In \emph{The Eleventh International Conference on Learning Representations}.

\end{thebibliography}

\clearpage

\appendix

\section{Appendix}
\label{sec:appendix}

\subsection{Prompt}
\label{app:prompt}
Prompt 1: \newline
<<question>>\newline
If you need to solve a current problem for a current problem, what relevant knowledge do you need? Ask a question about relevant knowledge. Please note: You only need to ask the question, you do not need to answer it.
\newline
\newline
Prompt 2:\newline
The answers should be short, but organized and informative.\newline
<<Step Guider Question>>
\newline
\newline
Prompt 3:\newline
If you need to solve the current problem for the current step, what relevant knowledge will be needed in the future?
\newline
\newline
Prompt 4:\newline
Next for the current topic to continue the next step of the answer, be sure not to repeat the previous content, to answer according to the previous content.\newline
<<Step Guidance>>

\subsection{Comparison}

As shown in Table \ref{app:tab:comparison} and Table \ref{app:tab:comparison_stem}, we show full comparison of  State-Of-The-Art (SOTA) across different model architectures and datasets.

Standard MATH benchmark: SGR achieves substantial improvements over baseline methods on MATH datasets. For Qwen2-7b-inst, SGR delivers a remarkable +9.2\% improvement over CoT (36.3\% vs 27.1\%), while for the larger Qwen2-72b-inst model, SGR achieves +10.9\% improvement (47.4\% vs 36.5\%). The most significant gains are observed with LLaMA3.1-8b-inst, where SGR outperforms CoT by +7.7\% (29.2\% vs 21.5\%). Notably, SGR consistently surpasses SBP across all model configurations, demonstrating its superior reasoning capabilities.

Competition-Level Mathematics: On prestigious mathematical competitions (MATH (level-5), AMC23, AIME24, OLY), our method shows competitive performance. While the QwQ-32B-Preview thinking model achieves the highest scores (73.7\% average), our SGR method with Qwen2-72b-inst reaches 47.4\%, significantly outperforming traditional CoT approaches and demonstrating strong reasoning abilities on expert-level problems.

Beyond MATH dataset (Table \ref{app:tab:comparison_stem}): Across diverse STEM subjects (Physics, Chemistry, Biology, Computer Science, Math, Engineering), SGR maintains consistent advantages . For Qwen2-7b-inst, SGR achieves 82.3\% accuracy (+17.4\% over CoT), while Qwen2-72b-inst reaches 90.9\% (+5.6\% improvement). The method shows particular strength in mathematical reasoning tasks, where SGR with LLaMA3.1-8b-inst achieves 85.9\% accuracy.

The results reveal that SGR benefits scale effectively with model size. Larger models (72B parameters) show more substantial absolute improvements, while smaller models (7B-8B parameters) demonstrate higher relative gains, indicating the method's broad applicability across different computational budgets.

\begin{table*}[!t]
\centering
\resizebox{0.75\linewidth}{!}{
\begin{adjustbox}{center}

\begin{tabular}{lcccccccccccc}
\toprule
\multicolumn{1}{c}{\multirow{2}{*}{}} &  \multicolumn{1}{c}{\multirow{2}{*}{Method}} & \multicolumn{6}{c}{\textbf{MATH}}                                 & \multicolumn{1}{c}{\multirow{2}{*}{\textbf{AMC23}}} & \multicolumn{1}{c}{\multirow{2}{*}{\textbf{AIME24}}}  & \multirow{2}{*}{\textbf{OLY}} & \multirow{2}{*}{\textbf{Average}} \\
\cmidrule{3-8}
 & & L1    & L2    & L3    & L4    & L5    & \multicolumn{1}{c}{Average} &  &  &  & \\
\midrule
\emph{Thinking Model} \\
\multicolumn{1}{l}{QwQ-32B-Preview}       &  &\textcolor{red}{97.5} & \textcolor{red}{96.4} & \textcolor{red}{95.4} & \textcolor{red}{91.8} & \textcolor{red}{84.9} & \textcolor{red}{92.2} & \textcolor{red}{85.0} & \textcolor{red}{50.0} & \textcolor{red}{67.4} & \textcolor{red}{73.7} \\
\midrule
\emph{Math-Specific Models} \\
\multicolumn{1}{l}{Qwen2-Math-7b-inst}       &  &93.1 & 87.2 & 82.6 & 72.4 & 52.0 & 73.8 & 62.5 & 13.3 & 34.1 & 45.9 \\
\multicolumn{1}{l}{Qwen2-Math-72b-inst}      &  & 95.0 & \textbf{94.1} & 90.5 & 83.7 & 67.7 & 83.9  & 60.0   & 20.0  &  42.5   & 51.7\\
\multicolumn{1}{l}{Qwen2.5-Math-7b-inst}     &  &  95.4 & 93.0 & 89.7 & 82.7 & 67.4 & 83.2 & 62.5  & 33.3      & 37.3 & 54.1 \\
\multicolumn{1}{l}{Qwen2.5-Math-72b-inst}    &  &  \textbf{96.3} & 93.5 & \textbf{90.9} & \textbf{84.9} & \textbf{73.3} & \textbf{85.7}  & \textbf{70.0}   & \textbf{43.3}   & \textbf{60.6}  & \textbf{65.5}  \\
\midrule

\emph{General Models} \\
\multirow{2}{*}{GPT-4o}   & CoT & \textbf{95.0} & \textbf{91.7} & \textbf{86.0} & \textbf{74.9} & \textbf{53.8} & \textbf{76.6} & \textbf{15.0} & \textbf{10.0} & \textbf{43.3} & \textbf{36.2} \\
 & SBP & \colorbox{lightred}{91.3} & \colorbox{lightred}{88.3} & \colorbox{lightred}{81.1} & \colorbox{lightred}{71.5} & \colorbox{lightred}{51.2} & \colorbox{lightred}{73.0} & \colorbox{lightred}{\textbf{15.0}} & \colorbox{lightred}{6.7} & \colorbox{lightred}{\textbf{43.3}} & \colorbox{lightred}{34.4}\textcolor{gray}{(-1.8)} \\
\midrule
\multirow{5}{*}{Qwen2-7b-inst}  & CoT & 85.1 & 73.4 & 65.2 & 52.4 & 37.1 & 57.8                      & 28.8    & 1.5       & 20.1    & 27.1                     \\
& SBP   & \colorbox{lightred}{84.2} & \colorbox{lightred}{71.8} & \colorbox{lightred}{64.1} & \colorbox{lightred}{52.1} & \colorbox{lightgreen}{38.4} & \colorbox{lightred}{57.5} & \colorbox{lightred}{22.5} & \colorbox{lightred}{0.0} & \colorbox{lightgreen}{27.3} & \colorbox{lightred}{26.8}\textcolor{gray}{(-0.3)}  \\
&L2M & \colorbox{lightred}{82.9} & \colorbox{lightred}{71.4} & \colorbox{lightred}{62.6} & \colorbox{lightred}{49.9} & \colorbox{lightred}{33.7} & \colorbox{lightred}{55.2} & \colorbox{lightred}{15.0} & \colorbox{lightgreen}{3.4} & \colorbox{lightred}{\textbf{34.7}} & \colorbox{lightred}{26.1}\textcolor{gray}{(-1.0)}\\
 & SGR  & \colorbox{lightgreen}{90.2} & \colorbox{lightgreen}{81.3} & \colorbox{lightgreen}{74.6} & \colorbox{lightgreen}{68.3} & \colorbox{lightgreen}{\textbf{58.6}} & \colorbox{lightgreen}{\textbf{71.4}}  & \colorbox{lightgreen}{38.8} & \colorbox{lightred}{1.5} & \colorbox{lightgreen}{33.3}  & \colorbox{lightgreen}{36.3}\textcolor{gray}{(+9.2)}  \\
& BoN@16 &  \colorbox{lightgreen}{91.5} & \colorbox{lightgreen}{84.6} & \colorbox{lightgreen}{76.4} & \colorbox{lightgreen}{62.7} & \colorbox{lightgreen}{40.3} & \colorbox{lightgreen}{66.4} & \colorbox{lightgreen}{46.3}  &  \colorbox{lightgreen}{5.0} & \colorbox{lightgreen}{31.7} & \colorbox{lightgreen}{37.4}\textcolor{gray}{(+10.3)}\\
& BoN@32 & \colorbox{lightgreen}{\textbf{92.8}} & \colorbox{lightgreen}{\textbf{85.5}} & \colorbox{lightgreen}{\textbf{79.7}} & \colorbox{lightgreen}{\textbf{66.8}} & \colorbox{lightgreen}{44.5} & \colorbox{lightgreen}{69.4} & \colorbox{lightgreen}{\textbf{52.5}}  & \colorbox{lightgreen}{\textbf{10.0}} & \colorbox{lightgreen}{34.4}  & \colorbox{lightgreen}{\textbf{41.6}}\textcolor{gray}{(+14.5)}\\
\midrule
\multirow{3}{*}{Qwen2-72b-inst} &  CoT &  91.4 & 85.3 & 77.3 & 66.9 & 46.1 & 69.2 & 35.0 & 6.0 & 35.8 & 36.5                    \\
& SBP & \colorbox{lightred}{88.6} & \colorbox{lightred}{82.2} & \colorbox{lightred}{72.1} & \colorbox{lightred}{60.2} & \colorbox{lightred}{38.7} & \colorbox{lightred}{63.6} & \colorbox{lightgreen}{36.3} & \colorbox{lightred}{1.7} & \colorbox{lightred}{32.7}  & \colorbox{lightred}{33.6}\textcolor{gray}{(-2.9)} \\
& L2M  &  \colorbox{lightgreen}{92.9} & \colorbox{lightgreen}{90.8} & \colorbox{lightgreen}{83.7} & \colorbox{lightgreen}{74.8} & \colorbox{lightgreen}{54.8} & \colorbox{lightgreen}{75.9}  &  \colorbox{lightgreen}{41.3} & \colorbox{lightgreen}{6.7} & \colorbox{lightgreen}{\textbf{44.0}} & \colorbox{lightgreen}{42.0}\textcolor{gray}{(+5.5)} \\
& SGR& \colorbox{lightgreen}{\textbf{93.9}} & \colorbox{lightgreen}{\textbf{89.3}} & \colorbox{lightgreen}{\textbf{83.7}} & \colorbox{lightgreen}{\textbf{76.9}} & \colorbox{lightgreen}{\textbf{65.6}} & \colorbox{lightgreen}{\textbf{79.2}}  & \colorbox{lightgreen}{\textbf{61.3}}    & \colorbox{lightgreen}{\textbf{8.0}} & \colorbox{lightgreen}{41.2} & \colorbox{lightgreen}{\textbf{47.4}}\textcolor{gray}{(+10.9)}                \\ \midrule
\multirow{3}{*}{LLaMA3.1-8b-inst} &  CoT &  76.2 & 61.2 & 50.8 & 36.6 & 21.2 & 43.7                      & \textbf{20.0}    & \textbf{8.0}       & 14.4         & 21.5                \\
& SBP & \colorbox{lightred}{75.3} & \colorbox{lightred}{59.3} & \colorbox{lightred}{48.1} & \colorbox{lightred}{36.4} & \colorbox{lightred}{21.2} & \colorbox{lightred}{42.5} & \colorbox{lightred}{11.3} & \colorbox{lightred}{5.0} & \colorbox{lightgreen}{18.5} & \colorbox{lightred}{19.3}\textcolor{gray}{(-2.2)} \\
& L2M & \colorbox{lightgreen}{85.2} & \colorbox{lightgreen}{72.4} & \colorbox{lightgreen}{62.4} & \colorbox{lightgreen}{48.7} & \colorbox{lightgreen}{31.6} & \colorbox{lightgreen}{54.7} & \colorbox{lightred}{17.5} & \colorbox{lightred}{5.0} & \colorbox{lightgreen}{\textbf{27.1}} & \colorbox{lightgreen}{26.1}\textcolor{gray}{(+4.6)} \\
& SGR &  \colorbox{lightgreen}{\textbf{81.7}} & \colorbox{lightgreen}{\textbf{76.8}} & \colorbox{lightgreen}{\textbf{71.5}} & \colorbox{lightgreen}{\textbf{66.8}} & \colorbox{lightgreen}{\textbf{61.2}} & \colorbox{lightgreen}{\textbf{69.5}} & \colorbox{lightred}{18.8} & \colorbox{lightred}{6.0} & \colorbox{lightgreen}{22.7}     & \colorbox{lightgreen}{\textbf{29.2}}\textcolor{gray}{(+7.7)}           
\\ 
\midrule
\multirow{3}{*}{LLaMA2-70b-inst} &  CoT & \textbf{44.5} & 25.4 & 15.8 & 9.6 & 5.2 & 15.7 & 4.0 & 0.0 & 2.3 & 5.5 \\
& SBP & \colorbox{lightred}{39.8} & \colorbox{lightgreen}{\textbf{26.1}} & \colorbox{lightgreen}{\textbf{19.1}} & \colorbox{lightgreen}{\textbf{14.8}} & \colorbox{lightgreen}{\textbf{14.7}} & \colorbox{lightgreen}{\textbf{19.9}} & \colorbox{lightgreen}{\textbf{6.3}} &\colorbox{lightred}{ 0.0} & \colorbox{lightgreen}{\textbf{5.1}} & \colorbox{lightgreen}{\textbf{7.8}}\textcolor{gray}{(+2.3)} \\
& SGR & \colorbox{lightred}{38.7} & \colorbox{lightred}{25.3} & \colorbox{lightgreen}{16.8} & \colorbox{lightgreen}{11.3} & \colorbox{lightgreen}{7.1} & \colorbox{lightgreen}{16.3} & \colorbox{lightgreen}{5.0} & \colorbox{lightgreen}{3.3} & \colorbox{lightgreen}{2.7} & \colorbox{lightgreen}{6.8}\textcolor{gray}{(+1.3)} \\  

\bottomrule
\end{tabular}

\end{adjustbox}
}


\caption{Accuracy comparison (\%) of CoT, SBP(5-shot) and our \our methods with the SOTA on MMLU-STEM dataset. We also report the results of open-sourced SOTA math-specific models - the QwQ, Qwen-Math models and GPT-4o. The best results are in \textbf{Bold} for each base and \textcolor{red}{Red} denotes the highest score in the current test set. \colorbox{lightred}{Red} indicates lower results compared to CoT, while \colorbox{lightgreen}{Green} denotes higher results.}
\label{app:tab:comparison}

\end{table*}

\begin{table*}[!t]
\centering
\resizebox{0.75\linewidth}{!}{
\begin{adjustbox}{center}

\begin{tabular}{lccccccccc}
\toprule
\multicolumn{1}{c}{\multirow{2}{*}{}} & \multicolumn{1}{c}{\multirow{2}{*}{Method}} & \multicolumn{7}{c}{\textbf{MMLU-STEM}}                                 \\
\cmidrule{3-9}
 &  & Physics    & Chemistry    & Biology    & Computer Science    & Math  & Engineer  & \multicolumn{1}{c}{Average}   \\
  \midrule
 \emph{Thinking Model} \\
 \multicolumn{1}{l}{QwQ-32B-Preview}  &  & \textcolor{red}{93.9} & 83.1 & 94.0 & 88.8 & \textcolor{red}{95.1} & \textcolor{red}{86.1} & \textcolor{red}{91.8} \\
 \midrule
\emph{Math-Specific Models} \\
\multicolumn{1}{l}{Qwen2-Math-7b-inst}       &  & 69.1 & 57.5 & 64.4 & 65.3 & 84.3 & 62.5 & 71.5 \\
\multicolumn{1}{l}{Qwen2-Math-72b-inst}      &   & 87.3 & 78.1 & \textbf{88.1} & 81.9 & 90.7 & 79.9 & 86.2    \\
\multicolumn{1}{l}{Qwen2.5-Math-7b-inst}     &   & 71.3 & 61.1 & 61.9 & 66.7 & 86.8 & 61.1 & 73.0 \\
\multicolumn{1}{l}{Qwen2.5-Math-72b-inst}    &   & \textbf{88.2} & \textbf{78.7} & 86.9 & \textbf{83.9} & \textbf{92.6} & \textbf{81.2} & \textbf{87.3} 
  \\
\midrule
\emph{General Models} \\
\multirow{2}{*}{GPT-4o}  & CoT & \textbf{90.0} & 64.8 & 94.7 & 85.3 & 87.8 & \textbf{83.3} & 86.1 \\
& SBP   & \colorbox{lightred}{89.6} & \colorbox{lightgreen}{\textbf{82.1}} & \colorbox{lightgreen}{\textbf{95.1}} & \colorbox{lightgreen}{\textbf{87.0}} & \colorbox{lightgreen}{\textbf{87.9}} & \colorbox{lightred}{77.8} & \colorbox{lightgreen}{\textbf{87.8}}\textcolor{gray}{(+1.7)} \\
\midrule
\multirow{5}{*}{Qwen2-7b-inst} & CoT & 65.9 & 56.0 & 79.5 & 64.7 & 73.2 & 62.2 & 64.9\\
& SBP  & \colorbox{lightred}{65.4} & \colorbox{lightred}{54.7} & \colorbox{lightred}{76.2} & \colorbox{lightgreen}{65.2} & \colorbox{lightred}{70.6} & \colorbox{lightgreen}{65.3} & \colorbox{lightgreen}{67.5}\textcolor{gray}{(+2.9)}\\
&L2M & \colorbox{lightred}{62.8} & \colorbox{lightred}{51.9} & \colorbox{lightred}{77.3} & \colorbox{lightgreen}{69.0} & \colorbox{lightred}{72.8}	 & \colorbox{lightgreen}{66.7} & \colorbox{lightgreen}{68.2}\textcolor{gray}{(+3.3)} \\
& SGR  & \colorbox{lightgreen}{\textbf{79.2}} & \colorbox{lightgreen}{\textbf{72.3}} &  \colorbox{lightgreen}{\textbf{88.9}} & \colorbox{lightgreen}{\textbf{85.2}} & \colorbox{lightgreen}{\textbf{84.1}} & \colorbox{lightgreen}{\textbf{74.0}} & \colorbox{lightgreen}{\textbf{82.3}}\textcolor{gray}{(+17.4)} \\
& BoN@16 & \colorbox{lightgreen}{67.9} & \colorbox{lightgreen}{56.1} & \colorbox{lightgreen}{80.0} & \colorbox{lightgreen}{66.1} & \colorbox{lightgreen}{82.1} & \colorbox{lightred}{59.7} & \colorbox{lightgreen}{73.0}\textcolor{gray}{(+8.1)}\\
& BoN@32   & \colorbox{lightgreen}{71.2} & \colorbox{lightgreen}{60.8} & \colorbox{lightgreen}{82.2} & \colorbox{lightgreen}{67.2} & \colorbox{lightgreen}{83.9} & \colorbox{lightred}{61.5} & \colorbox{lightgreen}{75.4}\textcolor{gray}{(+10.5)} \\
\midrule
\multirow{3}{*}{Qwen2-72b-inst}  & CoT & 86.3 & 74.9 & 93.8 & 81.8 & 86.5 & 75.3 & 85.3 \\
& SBP  & \colorbox{lightred}{81.8} & \colorbox{lightred}{70.6} & \colorbox{lightred}{91.4} & \colorbox{lightred}{80.3} & \colorbox{lightred}{82.7} & \colorbox{lightred}{71.9} & \colorbox{lightred}{81.5}\textcolor{gray}{(-3.8)} \\
&L2M & \colorbox{lightred}{80.8} & \colorbox{lightred}{71.9} & \colorbox{lightred}{89.7} & \colorbox{lightgreen}{82.8} & \colorbox{lightred}{86.5} & \colorbox{lightgreen}{76.8} & \colorbox{lightred}{83.0}\textcolor{gray}{(-2.3)} \\
& SGR  & \colorbox{lightgreen}{\textbf{90.7}} & \colorbox{lightgreen}{\textcolor{red}{\textbf{83.2}}} & \colorbox{lightgreen}{\textcolor{red}{\textbf{95.1}}} & \colorbox{lightgreen}{\textcolor{red}{\textbf{91.3}}} & \colorbox{lightgreen}{\textbf{92.7}} & \colorbox{lightgreen}{\textbf{78.8}} & \colorbox{lightgreen}{\textbf{90.9}}\textcolor{gray}{(+5.6)}                  \\
\midrule
\multirow{3}{*}{LLaMA3.1-8b-inst}  & CoT & 59.4 & 62.4 & 56.1 & 78.4 & 61.2 & 64.9 &  69.2 \\
& SBP   & \colorbox{lightgreen}{62.7} &  \colorbox{lightred}{57.7} & \colorbox{lightgreen}{77.6} &  \colorbox{lightred}{60.2} & \colorbox{lightgreen}{65.4} & \colorbox{lightgreen}{65.7} &  \colorbox{lightred}{64.9}\textcolor{gray}{(-4.3)} \\
&L2M  & \colorbox{lightgreen}{64.0} & \colorbox{lightred}{52.4} & \colorbox{lightgreen}{75.8} & \colorbox{lightred}{65.0} & \colorbox{lightgreen}{69.2} & \colorbox{lightred}{64.6} & \colorbox{lightred}{66.4}\textcolor{gray}{(-2.8)} \\
& SGR  & \colorbox{lightgreen}{\textbf{77.7}} & \colorbox{lightgreen}{\textbf{82.1}} & \colorbox{lightgreen}{\textbf{78.6}} & \colorbox{lightgreen}{\textbf{89.2}} & \colorbox{lightgreen}{\textbf{85.9}} & \colorbox{lightgreen}{\textbf{81.1}} & \colorbox{lightgreen}{\textbf{82.4}}\textcolor{gray}{(+13.2)}                 \\
\midrule
\multirow{3}{*}{LLaMA2-70b-inst}  & CoT & 46.0 & 39.4 & 72.0 & 55.9 & 38.7 & 51.8 & 48.1 \\
& SBP & \colorbox{lightgreen}{63.4} & \colorbox{lightgreen}{58.7} & \colorbox{lightgreen}{75.3} & \colorbox{lightgreen}{63.2} & \colorbox{lightgreen}{52.1} & \colorbox{lightgreen}{61.5} & \colorbox{lightgreen}{60.3}\textcolor{gray}{(+12.2)} \\
& SGR  & \colorbox{lightgreen}{\textbf{69.3}} & \colorbox{lightgreen}{\textbf{62.3}} & \colorbox{lightgreen}{\textbf{83.1}} & \colorbox{lightgreen}{\textbf{75.3}} & \colorbox{lightgreen}{\textbf{57.9}} & \colorbox{lightgreen}{\textbf{71.5}} & \colorbox{lightgreen}{\textbf{67.3}}\textcolor{gray}{(+19.2)}  \\

\bottomrule
\end{tabular}

\end{adjustbox}
}
\caption{Accuracy comparison (\%) of CoT, SBP(5-shot) and our \our methods with the SOTA over MATH (Level 1 to Level 5), AMC23, AIME24 and OLY datasets. We also report the results of open-sourced SOTA math-specific models - the QwQ, Qwen-Math models and GPT-4o. The best results are in \textbf{Bold} for each base and \textcolor{red}{Red} denotes the highest score in the current test set. \colorbox{lightred}{Red} indicates lower results compared to CoT, while \colorbox{lightgreen}{Green} denotes higher results.}
\label{app:tab:comparison_stem}

\end{table*}

\subsection{Dataset}
\label{app:dataset}

\begin{itemize}
    \item \textbf{MATH:}
    The MATH dataset comprises a substantial collection of 12,500 high school-level mathematical problems, meticulously curated to cover a wide range of topics and difficulty levels. In our study, we selected the MATH dataset's test data (5,000 problems) to evaluate our model's performance across diverse mathematical topics and difficulty levels, ensuring a robust assessment of its generalization and problem-solving capabilities.
    \item \textbf{AMC23:}
    It contains 40 data items, each including a question and an answer. 
    \item \textbf{AIME24:}
    The AIME24 test set is from the 2024 American Invitational Mathematics Examination. It has 30 questions, each with an answer. Among all our test sets, AIME24 is the most difficult. 
    \item \textbf{MMLU-STEM:}
    MMLU, or Massive Multitask Language Understanding, is a crucial benchmark for evaluating large language models. We have selected a test set with the MMLU-STEM label, which consists of a total of 3,018 problems.
    \item \textbf{Olympiadbench:}
    OlympiadBench is a bilingual and multimodal scientific evaluation dataset at the Olympiad level jointly, which contains 8,952 math and physics questions from international Olympiads, Chinese Olympiads, Chinese college entrance examinations, and mock exams. We have selected a dataset with the "OE\_TO\_maths\_en\_COMP" tag, and there are a total of 675 problems in this dataset. 
\end{itemize}

\subsection{A full Context of Example}
\label{app:sample}
\paragraph{Question:}

Paul is at a train station and is waiting for his train. He isn't sure how long he needs to wait, but he knows that the fourth train scheduled to arrive at the station is the one he needs to get on. The first train is scheduled to arrive in 10 minutes, and this train will stay in the station for 20 minutes. The second train is to arrive half an hour after the first train leaves the station, and this second train will stay in the station for a quarter of the amount of time that the first train stayed in the station. The third train is to arrive an hour after the second train leaves the station, and this third train is to leave the station immediately after it arrives.  The fourth train will arrive 20 minutes after the third train leaves, and this is the train Paul will board.  In total, how long, in minutes, will Paul wait for his train?

\paragraph{1-st iteration:}

    \begin{itemize}
        \item \textcolor{sgq}{Step Guidance Question}
        
        What is the schedule and duration of stay for each train arriving at the station before Paul's train, and how does this affect the total waiting time for Paul's train?
        
        \item \textcolor{sg}{Step Guidance}
        
        To answer your question, I would need more specific information about which station and which trains you are referring to. However, I can provide a general explanation of how train schedules and durations of stay can affect waiting times.

        The schedule of each train arriving at a station is typically determined by the train operator and can be found on their website or at the station. The duration of stay for each train, also known as the dwell time, is the time the train spends at the station for passengers to board and disembark. This can vary depending on the number of passengers and the type of train.

        If Paul's train is scheduled to arrive after several other trains, the total waiting time for his train may be affected by the dwell times of the previous trains. If the previous trains have longer dwell times, this can cause delays for subsequent trains, including Paul's train. However, if the previous trains have shorter dwell times and are on schedule, Paul's train may arrive on time or even early.

        Overall, the schedule and duration of stay for each train can have a significant impact on the waiting time for subsequent trains, and it's important for train operators to manage these factors to minimize delays and ensure on-time performance.

        \item Response for the First iteration
        
        To calculate the total waiting time for Paul, we need to consider the arrival and departure times of each train and the duration of stay for each train.

        1. The first train arrives in 10 minutes and stays for 20 minutes. Therefore, it will depart 30 minutes after its arrival.

        2. The second train arrives 30 minutes after the first train leaves. Since the first train leaves 30 minutes after its arrival, the second train arrives 60 minutes after the first train's arrival. The second train stays for a quarter of the time the first train stayed, which is 5 minutes (20 minutes / 4). Therefore, the second train will depart 65 minutes after the first train's arrival.
    \end{itemize}


\paragraph{2-nd iteration}
\begin{itemize}
\item \textcolor{sg}{Step Guidance}

3. The third train arrives an hour after the second train leaves. Since the second train leaves 65 minutes after the first train's arrival, the third train arrives 125 minutes after the first train's arrival (65 minutes + 60 minutes). The third train leaves immediately after it arrives, so it doesn't add any additional waiting time for Paul.

4. The fourth train, which is Paul's train, arrives 20 minutes after the third train leaves. Since the third train leaves immediately after it arrives, the fourth train arrives 145 minutes after the first train's arrival (125 minutes + 20 minutes).

To find out how long Paul waits for his train, we need to calculate the time from when he arrives at the station to when the fourth train arrives. Since the first train arrives in 10 minutes, Paul arrives at the station at this time. Therefore, Paul waits for his train for 135 minutes (145 minutes - 10 minutes).

In the future, to solve similar problems, you will need to know:
- The schedule of the trains, including arrival and departure times.
- The duration of stay for each train.
- The sequence of the trains that Paul needs to board.
- The time Paul arrives at the station.

With this information, you can calculate the total waiting time by determining the arrival time of the required train and subtracting the time Paul arrives at the station.
\item Response for the Second iteration
Given the information provided:

1. The first train arrives in 10 minutes and stays for 20 minutes, so it departs 30 minutes after its arrival.

2. The second train arrives 30 minutes after the first train leaves, which is 60 minutes after the first train's arrival. It stays for 5 minutes (a quarter of the first train's stay), so it departs 65 minutes after the first train's arrival.
    \end{itemize}


\paragraph{3-rd iteration}
\begin{itemize}
\item \textcolor{sg}{Step Guidance}

3. The third train arrives an hour after the second train leaves. Since the second train departs 65 minutes after the first train's arrival, the third train arrives 125 minutes after the first train's arrival (65 minutes + 60 minutes). The third train leaves immediately after it arrives, so it doesn't add any additional waiting time for Paul.

4. The fourth train, which is Paul's train, arrives 20 minutes after the third train leaves. Since the third train arrives and leaves immediately, the fourth train arrives 145 minutes after the first train's arrival (125 minutes + 20 minutes).

To find out how long Paul waits for his train, we need to calculate the time from when Paul arrives at the station to when the fourth train arrives. Since Paul arrives at the station at the same time the first train is scheduled to arrive (10 minutes from now), we can subtract the 10 minutes from the 145 minutes to find Paul's total waiting time.

Paul's total waiting time = 145 minutes - 10 minutes = 135 minutes.

Therefore, Paul will wait for 135 minutes for his train to arrive. The relevant knowledge needed for future steps would be the ability to calculate time intervals, understand schedules, and perform basic arithmetic operations. Additionally, understanding the concept of elapsed time and how to calculate it based on given schedules would be beneficial.

\item Response for the third iteration

Continuing from the previous steps:

3. The third train arrives an hour after the second train leaves. Since the second train departs 65 minutes after the first train's arrival, the third train arrives 125 minutes after the first train's arrival (65 minutes + 60 minutes). The third train leaves immediately after it arrives, so it doesn't add any additional waiting time for Paul.
    \end{itemize}

\paragraph{4-th iteration}
\begin{itemize}
\item \textcolor{sg}{Step Guidance}

4. The fourth train, which is Paul's train, arrives 20 minutes after the third train leaves. Since the third train arrives and leaves at the same time, which is 125 minutes after the first train's arrival, the fourth train arrives 145 minutes after the first train's arrival.

To find out how long Paul waits for his train, we need to consider when he arrives at the station. Since the problem doesn't specify when Paul arrives, we can assume he arrives just before the first train is scheduled to arrive, which is at the 10-minute mark. Therefore, Paul's waiting time is the time from his arrival until the fourth train arrives.

Given that Paul arrives just before the first train arrives (at the 10-minute mark) and the fourth train arrives 145 minutes after the first train's arrival, Paul's total waiting time is 145 minutes.

However, since Paul arrives 10 minutes before the first train, we need to subtract those 10 minutes from the 145 minutes to find the total time Paul actually waits from his arrival. Therefore, Paul's total waiting time is 135 minutes (145 minutes - 10 minutes).

In summary, the relevant knowledge needed to solve this problem includes understanding the schedule of each train, calculating the departure times based on arrival and stay times, and determining the total waiting time for Paul based on his arrival time and the arrival time of his train.

\item Response for the Fourth iteration

Given the sequence of events and the timing for each train's arrival and departure:

1. The first train arrives 10 minutes after Paul's arrival and stays for 20 minutes, departing 30 minutes after Paul's arrival.
2. The second train arrives 30 minutes after the first train leaves, which is 60 minutes after Paul's arrival, and stays for 5 minutes, departing 65 minutes after Paul's arrival.
3. The third train arrives an hour after the second train leaves, which is 125 minutes after Paul's arrival, and departs immediately.
\end{itemize}
\paragraph{5-th iteration}
\begin{itemize}
\item \textcolor{sg}{Step Guidance}

To find out how long Paul will wait for his train, we need to calculate the arrival time of the fourth train and compare it to the time Paul arrived at the station. Here's the breakdown:

1. **First Train:**
   - Arrives 10 minutes after Paul's arrival.
   - Stays for 20 minutes.
   - Departs 30 minutes after Paul's arrival.

2. **Second Train:**
   - Arrives 30 minutes after the first train leaves, which is 60 minutes after Paul's arrival.
   - Stays for 5 minutes (a quarter of the first train's stay).
   - Departs 65 minutes after Paul's arrival.

3. **Third Train:**
   - Arrives an hour after the second train leaves.
   - Since the second train departs 65 minutes after Paul's arrival, the third train arrives 125 minutes after Paul's arrival (65 minutes + 60 minutes).
   - Leaves immediately after it arrives.

4. **Fourth Train:**
   - Arrives 20 minutes after the third train leaves.
   - Since the third train arrives 125 minutes after Paul's arrival and leaves immediately, the fourth train arrives 145 minutes after Paul's arrival.

Given that Paul arrived at the station at time 0, the total waiting time for Paul is the time the fourth train arrives, which is 145 minutes. Therefore, Paul will wait for 145 minutes for his train to arrive. 

This calculation takes into account the arrival and departure times of each train and the duration of stay for each train, as well as the time between the departure of one train and the arrival of the next. This information is necessary to determine the exact waiting time for Paul.
\item Response for the Fifth iteration

Continuing from the established sequence:

4. The fourth train, which is the one Paul is waiting for, arrives 20 minutes after the third train leaves. Since the third train arrives 125 minutes after Paul's arrival and departs immediately, the fourth train arrives 145 minutes after Paul's arrival.

Therefore, Paul's total waiting time is 145 minutes, which is the time it takes for the fourth train to arrive after he initially arrives at the station. This total includes the waiting times for all the preceding trains to arrive, depart, and the intervals between their departures and the arrivals of the subsequent trains.
\end{itemize}
\subsection{Comparison with CoT and \our}
\label{app:r1}
\begin{figure*}[!t]
    \centering
    \includegraphics[width=1.00\linewidth]{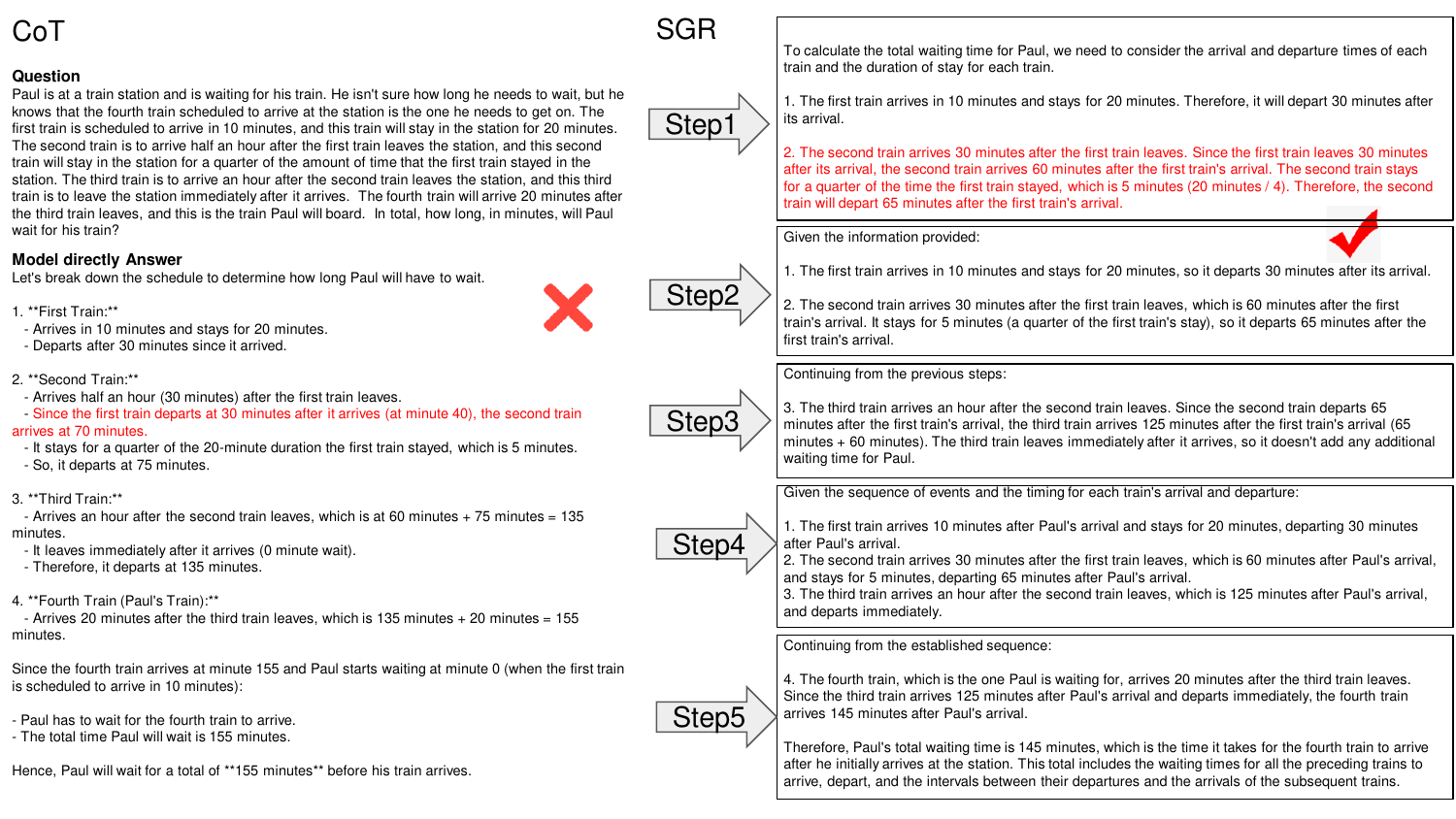}
    \caption{The comparison above shows the results of models using direct answering versus the \our approach. The red sections in the direct answers indicate errors, while the corresponding red sections in the \our answers are correct. Each step of the \our-generated answer is enclosed in a box.}
    \label{fig:sample}
\end{figure*}

\end{document}